\documentclass[letterpaper]{article} 
\usepackage[preprint]{aaai2027}  
\usepackage[hyphens]{url}  
\usepackage{graphicx} 
\usepackage{natbib}  
\usepackage{caption} 
\usepackage{algorithm}
\usepackage{algorithmic}

\usepackage{newfloat}
\usepackage{listings}
\DeclareCaptionStyle{ruled}{labelfont=normalfont,labelsep=colon,strut=off} 
\floatstyle{ruled}
\newfloat{listing}{tb}{lst}{}
\floatname{listing}{Listing}

\usepackage{booktabs}

\usepackage{amsmath}
\usepackage{amsfonts}
\usepackage{amssymb}
\usepackage{amsthm}
\usepackage{multirow}
\usepackage{thm-restate}

\newcommand{\method}{PD-NUUS}

\theoremstyle{plain}

\newtheorem{lemma}{Lemma}

\title{Beyond $L_\infty$-Norm Sets: Synthesizing Interpretable Non-Uniform Uncertainty Sets for Evaluation-Time State-Adversarial Robustness}
\author{
    Xulin Chen,
    Ruipeng Liu,
    Garrett E. Katz
}
\affiliations{
    Department of Electrical Engineering and Computer Science \\
    Syracuse University \\

    Syracuse, NY 13244 USA\\
    \{xchen168,rliu02,gkatz01\}@syr.edu
}

\begin{document}

\maketitle

\begin{abstract}
Deep reinforcement learning (DRL) agents are vulnerable to perturbations of their observed state inputs. Prior works often constrain state adversaries with $L_p$-norm uncertainty sets and evaluate policy performance under the resulting perturbations. However, in continuous control tasks like robotics control, different state dimensions have distinct physical meanings and influence policy behavior to different degrees, which $L_p$-norm sets fail to capture or explain. Moreover, it is important to identify when and along which state dimensions a policy is vulnerable, independently of the algorithm used to train it. In this paper, we focus on evaluation-time attacks applied to pretrained policies. We propose \textbf{P}erformance-\textbf{D}egradation \textbf{N}on-\textbf{U}niform \textbf{U}ncertainty \textbf{S}et (\method{}), an interpretable formulation of uncertainty sets for state perturbations. \method{} is state-dependent, dimension-wise non-uniform, and can be synthesized online during policy evaluation. We first formulate an ideal constrained optimization problem for \method{} computation, then introduce a surrogate constraint to make the problem tractable. Experiments across multiple continuous-control environments, policy algorithms, and state attacks show that \method{} reveals variation in policy sensitivity across states and dimensions, while its performance degradation is moderately controlled.
\end{abstract}


\section{Introduction}
Deep reinforcement learning (DRL) has achieved remarkable success in complex control problems, including robotics and autonomous driving. However, even well-trained policies may remain vulnerable to perturbations in observed states, transition dynamics, or action execution, which substantially degrade policy performance and cause unsafe behavior. Improving policy robustness against these perturbations is therefore essential when deploying DRL policies.

In this work, we focus on policy robustness against observed state perturbations in continuous control tasks. While some studies consider more structured formulations \citep{mandlekar2017adversarially,everett2021certifiable}, prior methods commonly constrain perturbations using an $L_\infty$-norm uncertainty set which assumes a fixed perturbation budget across all states and state dimensions, and then evaluate or optimize the resulting performance degradation
\citep{zhang2020robust,oikarinen2021robust,zhang2021robust,liang2022efficient}. Although convenient, the $L_\infty$-norm formulation has two important limitations that hinder its interpretability. First, a fixed perturbation budget does not account for variations in policy sensitivity across states and state dimensions. State features can differ substantially in their physical meanings and contributions to policy behavior (see a toy experiment in Table \ref{tab:contribution-of-state-dim-to-policy-performance}). Even perturbations along the same dimension have different effects depending on the current state; for example, a robot is more sensitive to perturbations when it is about to fall than when it is stable. Therefore, $L_\infty$-norm produces insufficiently realistic perturbations that are restrictive in some scenarios but damaging in others. Second, in prior work the perturbation budget is often selected empirically and lacks a direct interpretation in terms of performance degradation. An appropriate budget for one environment or policy-learning algorithm can be unsuitable for another. While some works \citep{zhang2020robust,zhang2021robust} show how to design adversarial perturbations within a specified uncertainty set, they rarely explore how to design a policy-specific uncertainty set which controls the resulting performance degradation.

\begin{table*}[t]
    \centering
    \fontsize{9pt}{11pt}\selectfont

    {
    \renewcommand{\arraystretch}{1.1}
    
    

    \begin{tabular}{cccccccccccc}
        \hline
        \multirow{3}{*}{\shortstack{\textbf{Noise}\\($\delta$)}}
        & \multicolumn{11}{c}{\textbf{State Dimensions}} \\
        \cmidrule{2-12}
    
        & \multirow{2}{*}{\shortstack{Hopper\\Height}}
        & \multirow{2}{*}{\shortstack{Torso\\Angle}}
        & \multicolumn{3}{c}{\shortstack{Joint\\Angles}}
        & \multicolumn{2}{c}{\shortstack{Torso Linear\\Velocities}}
        & \multirow{2}{*}{\shortstack{Torso Angular\\Velocity}}
        & \multicolumn{3}{c}{\shortstack{Joint\\Velocities}} \\
    
        \cmidrule(lr){4-6}
        \cmidrule(lr){7-8}
        \cmidrule(lr){10-12}
    
        & & &
        Thigh & Leg & Foot &
        X-Axis & Z-Axis &
        &
        Thigh & Leg & Foot \\
        \hline
    
        0
        & \multicolumn{11}{c}{3547.9} \\
        \hline
    
        0.1
        & \underline{3282.5}
        & \textbf{3580.5}
        & 3486.3 & 3497.9 & 3499.9
        & 3490.9 & 3435.4
        & 3475.6
        & 3579.6 & 3561.0 & 3580.4 \\
    
        0.2
        & \underline{2169.2}
        & \textbf{3575.7}
        & 3440.7 & 3540.5 & 3294.4
        & 3245.6 & 3035.1
        & 3395.5
        & 3484.2 & 3472.8 & 2966.1 \\
    
        0.3
        & \underline{1501.0}
        & \textbf{3468.9}
        & 2918.6 & 3153.6 & 2632.2
        & 3217.5 & 1967.7
        & 1627.0
        & 3373.2 & 3342.9 & 2523.7 \\
    
        0.4
        & 1093.8
        & \textbf{3583.8}
        & 2172.6 & 3098.3 & 2539.1
        & 2399.4 & 994.7
        & \underline{414.6}
        & 2895.7 & 2896.6 & 1744.4 \\
        \hline
    \end{tabular}
    
    }
    \caption{A toy experiment in which each state dimension is perturbed for a PPO \citep{schulman2017proximal} policy trained on Hopper environment. To perturb the $i$-th state dimension, we apply constant noise, i.e. $s_i \leftarrow s_i + \delta$. We report the mean total reward over 20 episodes. For each value of noise $\delta$, the most and least robust dimensions are shown in \textbf{bold} and \underline{underlined}, respectively.}
    \label{tab:contribution-of-state-dim-to-policy-performance}
\end{table*}

The limitations motivate an interpretable formulation of uncertainty sets. We aim to answer the following question:
\begin{center}
    \textit{Given a policy, what is the largest uncertainty set whose admissible perturbations remain consistent with a predefined tolerance on performance degradation?}
\end{center}
Our main contributions are summarized as follows:
\begin{itemize}
    \item We propose \textbf{P}erformance-\textbf{D}egradation \textbf{N}on-\textbf{U}niform \textbf{U}ncertainty \textbf{S}et (\method{}), a state-dependent and non-uniform uncertainty set for observed state perturbations. Unlike $L_\infty$-norm, \method{} relates local perturbations to a predefined tolerance on degradation in discounted return, providing a performance-based interpretation of the admissible perturbations.
    \item We formulate the synthesis of \method{} as a constrained optimization problem, and derive a tractable surrogate constraint that mitigates occupancy shift with a policy divergence regularizer. The resulting closed-form solution can be directly applied to pretrained policies for robustness evaluation without changing the policy update rule.

    \item Extensive experiments across multiple policy-learning algorithms, environments, and state-adversarial attacks demonstrate that \method{} provides generally reliable performance-level control over the degradation in discounted return.
\end{itemize}

\section{Related Work}
\subsection{$L_p$-Norm Uncertainty Set in Robust RL}
Robust reinforcement learning (robust RL) studies how policies maintain reliable performance under perturbations to different components of a Markov Decision Process (MDP), including observed states and transition dynamics \cite{iyengar2005robust,zhang2020robust}. A single type of $L_p$-norm uncertainty set is most widely used to model perturbations. For example, \citet{zhang2020robust,oikarinen2021robust,zhang2021robust,liang2022efficient} improve policy robustness against state perturbations constrained within a predefined $L_\infty$-norm set. \citet{pattanaik2017robust} constrain adversarial state noise using an $L_2$-norm set. \citet{kuang2022learning} propose an $L_\infty$-norm based method to handle dynamics perturbations. Randomized smoothing approaches provide certified policy robustness under $L_2$-norm perturbations, although the certification is for a smoothed policy rather than the original policy \citep{kumar2022policy,mu2024reward}.


\subsection{Non-Uniform Perturbations}
\citet{erdemir2021adversarial} introduce non-uniform perturbations for classification and prediction problems, and \citet{wong2020learning} propose to learn a non-uniform perturbation set with conditional variational autoencoders (CVAEs). For vision-based RL, \citet{sun2026diffusion} employ diffusion models to generate realistic perturbations semantically distinct from the original images. In robust MDPs where transition dynamics are perturbed, \citet{petrik2019beyond,zhang2023robust} optimize the shape of the uncertainty set using value functions, while \citet{behzadian2021optimizing} leverage a percentile-performance criterion. For observed state perturbations, \citet{mandlekar2017adversarially} use the fast gradient method to create adversarial perturbations, and \citet{everett2021certifiable} certify the robustness of action-value function. Additionally, \citet{korkmaz2023detecting} identify adversarial directions through local quadratic approximations of policy loss, \citet{shen2020deep} improve robustness through policy smoothness regularization, and \citet{wu2021crop} certify action and cumulative-reward robustness through functional smoothing. However, we instead aim to explore how to attack a pretrained policy in a controlled way, and our method constrains the degradation directly in return rather than value functions, motivated by the observation from \citet{beechey2023explaining} that contributions of state features to the value function do not necessarily provide faithful insight into contributions to policy performance.

\section{Preliminaries}
\subsection{MDP and SA-MDP} We consider tasks with continuous state and action spaces. An MDP is specified by a tuple $(\mathcal{S}, \mathcal{A}, P,\rho,R,\gamma)$, where $\mathcal{S} \subseteq \mathbb{R}^m$ and $\mathcal{A}\subseteq  \mathcal{R}^n$ denote state and action spaces respectively, $P:\mathcal{S}\times \mathcal{A}\to \Delta_\mathcal{S}$ is the transition kernel with $\Delta_{(\cdot)}$ denoting the set of probability distributions over the given space, $\rho \in \Delta_{\mathcal{S}}$ is the initial state distribution, $R:\mathcal{S}\times \mathcal{A} \to \mathbb{R}$ is the reward function, and $\gamma \in (0,1)$ is the discount factor. A policy $\pi:\mathcal{S}\to\Delta_\mathcal{A}$ maps each state to a distribution over actions. The discounted return of $\pi$ is defined as
\begin{equation}
J(\pi) \triangleq \mathbb{E}_{s_0\sim \rho,a_t\sim \pi(\cdot|s_t), s_{t+1}\sim P(\cdot|s_t,a_t)}
\left[
\sum_{t=0}^{\infty}\gamma^t R(s_t,a_t)
\right].
\end{equation}
The advantage function $A^\pi(s,a)$ quantifies how much better a specific action is compared to the average action in that state. The normalized discounted future state distribution $d^\pi$ induced by $\pi$, also called the occupancy distribution, is defined as
\begin{equation}
d^\pi(s) \triangleq (1-\gamma)\sum_{t=0}^{\infty}\gamma^t
\Pr(s_t=s;\pi,\rho).
\label{eq:occupancy-definition}
\end{equation}
In this work, we focus on the scenarios in which the policy observes perturbed states. The State-Adversarial Markov Decision Process (SA-MDP) \citep{zhang2020robust} augments the standard MDP with a state adversary $\nu: \mathcal{S} \to \mathcal{S}$ only for the observed states. The action is predicted by a perturbed policy, while the environment still transitions from the true state $s$ to the next state. Although SA-MDP shows the influence of a state adversary on the Bellman operator during policy training, we only consider policy evaluation.

\subsection{$L_p$-Norm Uncertainty Set of State Adversary} 
The state adversary $\nu$ is typically constrained to make the problem well-posed. A common choice is an $L_p$-norm ball around each state $s \in \mathcal{S}$: That is, $\nu(s)\in B(s)$, where $B(s) \triangleq \{ \hat{s}\in\mathcal{S}| \ \ ||s-\hat{s}||_p \leq \epsilon \}$, and $\epsilon > 0$ controls the perturbation budget. Different choices of $p$-norm correspond to different assumptions on the perturbation geometry. A policy observes a perturbed state $\nu(s)$ sampled from $B(s)$ and uses $\nu(s)$ as the basis for selecting an action, though the actual state transition proceeds from the true state $s$.

\section{Methodology}





\subsection{Problem Formulation of \method{}}

We consider the scenario where a policy is evaluated under observed state perturbations. At each time step, if the policy reaches a state $s$, then a \method{} is synthesized online around $s$ after which a perturbation is sampled. Given a continuous state space $\mathcal{S}\subseteq\mathbb{R}^m$, we define \method{} with a state-dependent budget function $r:\mathcal{S} \to [0,r_{\max}]^m$
\begin{gather}
    B_r(s) \triangleq \left\{ \hat{s}\in\mathcal{S}| \ \ |s_i-\hat{s}_i| \leq r_i(s), \ i=1,\dots,m \right\}.
    \label{eq:nuus-definition}
\end{gather}
In other words, a \method{} $B_r(s)$ augments an $L_\infty$-norm ball with anisotropic scaling. We leave the correlation of state dimensions for future work. The class of admissible state adversaries induced by a \method{} is denoted as 
\begin{equation}
    \mathcal{V}_r \triangleq \{\nu:\mathcal{S}\rightarrow\mathcal{S} | \ \nu(s)\in B_r(s), \ \forall s\in\mathcal{S} \},
\end{equation}
and attacking the observed state of a clean policy $\pi$ with a state adversary $\nu \in \mathcal{V}_r$ induces a new policy $\pi_\nu(\cdot|s) =\pi(\cdot|\nu(s))$. The synthesis of \method{} is then formulated as a constrained functional optimization problem over $r(\cdot)$.
\subsubsection{Uncertainty Set Size as Objective.} 
Since $r_i(s)\geq0$ for all state dimensions, we define the objective $\Phi(\pi,r)$ as the expected 1-norm of $r(s)$ over states visited by the policy $\pi$:
\begin{equation}
    \Phi(\pi,r) = \mathbb{E}_{s\sim d^{\pi}} \left[ ||r(s)||_1 \right] = \mathbb{E}_{s\sim d^{\pi}}[\textbf{1}^\top r(s)]. \label{eq:nuus-objective-function}
\end{equation}

\subsubsection{Performance Degradation as Constraint.} We constrain the worst-case performance degradation in the expected discounted return. Given a tolerance $\beta>0$, all admissible adversaries are required to satisfy 
\begin{equation}
\sup_{\nu\in\mathcal{V}_r}
\left[ J(\pi)-J(\pi_\nu) \right] \leq \beta. \label{eq:ideal-performance-degradation-constraint}
\end{equation}
We apply the following performance difference equation which is obtained immediately by setting $\pi'=\pi_\nu$ in Lemma 6.1 of \citet{kakade2002approximately}:
\begin{equation}
    J(\pi) - J(\pi_\nu)
    = -\frac{1}{1-\gamma} \mathbb{E}_{s\sim d^{\pi_\nu},\,a\sim \pi_\nu} \left[ A^\pi(s,a) \right].
    \label{eq:pdl-for-state-adversary}
\end{equation}
Therefore, the constraint is equivalent to
\begin{equation}
     \sup_{\nu \in \mathcal{V}_r} -\mathbb{E}_{s\sim d^{\pi_\nu}, a\sim \pi_\nu}[A^\pi(s,a)] \leq (1-\gamma)\beta. \label{eq:nuus-inequality-constraint}
\end{equation}

\subsubsection{Ideal Problem Formulation.} Combining objective and constraint, the formal \method{} synthesis problem is:
\begin{align}
    \max_{r} & \quad \mathbb{E}_{s\sim d^{\pi}}[\textbf{1}^\top r(s)], \label{eq:nuus-ideal-problem-formulation-problem} \\
    \text{s.t.} & \quad  \sup_{\nu \in \mathcal{V}_r} -\mathbb{E}_{s\sim d^{\pi_\nu}, a\sim \pi_\nu}[A^\pi(s,a)] \leq (1-\gamma)\beta. \label{eq:nuus-ideal-problem-formulation-constraint}
\end{align}
Ideally, after specifying the hyperparameter $\beta$, the optimal solution $r^*(s)$ defines \method{} over the visited states such that the worst-case degradation in discounted total return does not exceed $\beta$. However, directly solving this problem is generally intractable. This is because the constraint contains a state adversary $\nu$, which has no explicit expression but is only known to be constrained. Although $\pi_\nu$ can be approximated through a Taylor expansion, accurately computing $d^{\pi_\nu}$ generally requires both knowledge of the environment dynamics and an explicitly specified $\pi_\nu$. Alternatively, $d^{\pi_\nu}$ may be estimated from trajectories sampled by $\pi_\nu$, but repeating this process throughout the optimization is computationally expensive. Moreover, unlike prior works on policy optimization \cite{schulman2015trust,achiam2017constrained,schulman2017proximal} which typically assume $d^{\pi_\nu} \approx d^\pi$ to guarantee policy improvement within constrained policy change, our formulation admits substantially perturbed policies, especially when $\beta$ is large. This implies $d^{\pi_\nu}$ may differ considerably from $d^\pi$.



\subsection{Approximate Closed-Form Solution}

\subsubsection{Surrogate Constraint for Intractable $d^{\pi_\nu}$.} Inspired by the divergence regularizer in offline RL \cite{wu2019behavior}, we replace the LHS of Eq. \eqref{eq:nuus-ideal-problem-formulation-constraint} by a surrogate
\begin{equation}
\sup_{\nu \in \mathcal{V}_r} -\mathbb{E}_{s\sim d^{\pi}} [\mathbb{E}_{a\sim \pi_\nu} [A^\pi(s,a)] - \alpha D_{\text{KL}} (\pi_\nu||\pi)[s] ],
    \label{eq:nuus-surrogate-constraint}
\end{equation}
where a constant $\alpha>0$ controls the strength of the Kullback–Leibler (KL) divergence penalty. In offline RL, such regularization handles occupancy shift by encouraging the learned policy to locally stay close to the behavior policy. In our setting, we use this approach to penalize state adversaries that induce large deviations between $\pi_\nu$ and $\pi$. The following lemma motivates the calibration of $\alpha$.
\begin{lemma}[Proof in Appendix~\ref{proof:occupancy-shift-on-nuus}]
\label{lemma:occupancy-shift-on-nuus}
Assume that a transition kernel is $L_P$-Lipschitz continuous w.r.t the action: $||P(\cdot|s,a_1)-P(\cdot|s,a_2)||_1\leq L_P ||a_1-a_2||_1$ for all $s\in \mathcal{S}$ and for all $a_1,a_2 \in \mathcal{A}$, and a stochastic policy $\pi$ is $L_\pi$-Lipschitz continuous from states to action distributions: $W_1(\pi(\cdot|s_1),\pi(\cdot|s_2)) \leq L_\pi ||s_1-s_2||_\infty$ for all $s_1,s_2\in \mathcal{S}$ where $W_1$ is the 1-Wasserstein distance induced by the $L_1$-norm on the action space $\mathcal{A}$. Then for any state adversary $\nu \in \mathcal{V}_r$,
\begin{equation}
    D_{\text{TV}}(d^{\pi},d^{\pi_\nu}) \leq \frac{\gamma L_PL_\pi r_{\max}}{2(1-\gamma)}.
\end{equation}
\end{lemma}
Although $D_{\text{TV}}(d^{\pi},d^{\pi_\nu})$ is over the entire state space and the Lipschitz assumption on transition dynamics is often violated in continuous control tasks, Lemma \ref{lemma:occupancy-shift-on-nuus} indicates that the occupancy shift can depend on both the policy sensitivity and environment dynamics. This observation motivates calibrating $\alpha$ separately for each environment and policy algorithm. In practice, we determine $\alpha$ empirically according to diagnostics obtained from trajectories sampled by $\pi$, and keep it fixed throughout the synthesis and evaluation of \method{}. We also provide an $O(\alpha^{-1})$ upper bound on the possible violation of the prescribed degradation tolerance $\beta$. Increasing $\alpha$ reduces this upper bound, although it does not necessarily make the actual degradation decrease monotonically.
\begin{lemma}[Proof in Appendix~\ref{proof:degradation-upperbound-on-nuus}]
\label{lemma:degradation-upperbound-on-nuus}
Consider a fixed policy $\pi$, $\gamma\in(0,1)$, and $\alpha,\beta>0$. After using the surrogate constraint in Eq. \eqref{eq:nuus-surrogate-constraint}, the supremum degradation of discounted return satisfies the following upper bound
\begin{equation}
\sup_{\nu\in\mathcal V_{r}}
\bigl[J(\pi)-J(\pi_\nu)\bigr]-\beta
\le
\frac{\gamma^2(\max_{s,a} |A^\pi(s,a)|)^2}{2\alpha(1-\gamma)^3}.
\end{equation}
\end{lemma}

\subsubsection{Reformulation with a Convex Quadratic Constraint.} 
\label{sec:convert-problem-into-qcqp-and-derive-solution}
The detailed derivation is provided in Appendix~\ref{sec:appendix-derive-solution}. To express Eq. \eqref{eq:nuus-surrogate-constraint} more directly in terms of $r(s)$, we approximate the perturbed policy $\pi_\nu$ using a second-order Taylor expansion, and approximate the local KL divergence $D_{\text{KL}}(\pi_\nu||\pi)[s]$ using the state Fisher information matrix. Letting $\delta(s) = \nu(s) - s$, the resulting supremum over $\nu \in \mathcal{V}_r$ becomes
\begin{equation}
    \sup_{\delta:|\delta_i(s)|\leq r_i(s)} \left\{\frac{1}{2}\delta(s)^\top H_{\pi,\alpha}(s) \delta(s) -g_\pi(s)^\top \delta(s) \right\}, \label{eq:supremum-of-surrogate-constraint}
\end{equation}
where
\begin{align}
& H_{\pi,\alpha}(s) = \alpha \mathbb{E}_{a\sim \pi}[\nabla_s \log \pi(a|s) \nabla_s \log \pi(a|s)^\top] \nonumber \\
& \quad -\mathbb{E}_{a\sim \pi} \Big[ A^\pi(s,a) \Big( \nabla_s \log\pi(a|s) \nabla_s \log\pi(a|s)^\top \nonumber \\
& \qquad \qquad + \nabla_s^2 \log\pi(a|s) \Big) \Big], \label{eq:nuus-reformulation-H-matrix}
\end{align}
and
\begin{equation}
    g_\pi(s) = \mathbb{E}_{a\sim \pi}[A^\pi(s,a) \nabla_s \log \pi(a|s)] \label{eq:nuus-h_pi}.
\end{equation}
The matrix $H_{\pi,\alpha}(s)$ is symmetric but not necessarily positive semidefinite. Consequently, Eq. \eqref{eq:supremum-of-surrogate-constraint} does not generally admit a simple analytical solution. To obtain a closed-form solution, we construct a diagonal positive-definite matrix $M_{\pi,\alpha}(s)$ which upper bounds $H_{\pi,\alpha}(s)$ in the Loewner order, meaning $M_{\pi,\alpha}(s) \succeq H_{\pi,\alpha}(s)$. This indicates that $\delta(s)^\top M_{\pi,\alpha}(s)\delta(s) \geq \delta(s)^\top H_{\pi,\alpha}(s)\delta(s)$ holds for every $\delta(s)$. We implement $M_{\pi,\alpha}(s)$ as a separable quadratic surrogate (SQS) majorization \cite{mcgaffin2015algorithmic}:
\begin{equation}
    M_{\pi,\alpha}(s) = \text{diag}_j \Big\{ \sum_{i=1}^m \big|[H_{\pi,\alpha}(s)]_{ij}\big| \Big\} + \tau I,
    \label{eq:separate-quadratic-surrogate-majorization}
\end{equation}
where $\tau>0$ is a small offset. Since $M_{\pi,\alpha}(s)$ is diagonal with positive diagonal entries, the majorized supremum is separable across dimensions and can be evaluated exactly:
\begin{align}
&\sup_{\delta:|\delta_i(s)|\leq r_i(s)} \left\{\frac{1}{2}\delta(s)^\top H_{\pi,\alpha}(s) \delta(s) -g_\pi(s)^\top \delta(s) \right\} \nonumber \\
&\leq \frac{1}{2}r(s)^\top M_{\pi,\alpha}(s) r(s) + |g_\pi(s)|^\top r(s). \label{eq:upper-bound-of-supremum-in-surrogate-constraint}
\end{align}
We replace the supremum in the surrogate constraint with the upper bound in Eq~\eqref{eq:upper-bound-of-supremum-in-surrogate-constraint}. Although the majorization induces a more conservative solution for $r(\cdot)$, it renders the constraint convex and analytically tractable. Finally, we obtain a tractable optimization problem for \method{} with $\beta>0$:
\begin{align}
\max_r & \quad  \mathbb{E}_{s\sim d^\pi} [ \textbf{1}^\top r(s)], \\ 
\text{s.t.} & \quad \mathbb{E}_{s\sim d^{\pi}}\Big[  \frac{1}{2}r(s)^\top M_{\pi,\alpha}(s) r(s) \nonumber \\
& \qquad \qquad \qquad + |g_\pi(s)|^\top r(s)  \Big] \leq (1-\gamma)\beta, \\
& \quad 0 \leq r_i(s) \leq r_{\max}, \ \ \forall s \in \mathcal{S}, \ i=1,\dots,m.
\label{eq:nuus-final-tractable-problem}
\end{align}
When the optimal solution $r^*_i(s)$ satisfies $0<r^*_i(s)<r_{\max}$ for every state dimension $i$, the box constraints are inactive. By complementary slackness, their associate Karush–Kuhn–Tucker (KKT) multipliers vanish and the stationarity condition yields a closed-form solution for $\beta>0$
\begin{equation}
r^*(s)=M_{\pi,\alpha}(s)^{-1} \left(\kappa \textbf{1} - |g_\pi(s)| \right), 
\label{eq:nuus-final-solution}
\end{equation}
where
\begin{equation}
\kappa= \sqrt{\frac{2(1-\gamma)\beta + \mathbb{E}_{s'\sim d^\pi} \left[ |g_\pi(s')|^\top M_{\pi,\alpha}(s')^{-1}|g_\pi(s')| \right]}{\mathbb{E}_{s'\sim d^\pi} \left[ \textbf{1}^\top M_{\pi,\alpha}(s')^{-1}\textbf{1} \right]}}
\label{eq:nuus-solution-kappa-term}
\end{equation}
is a global scalar estimated from trajectories collected under the clean policy $\pi$, and $g_\pi(s)$ and $M_{\pi,\alpha}(s)$ are defined in Eq. \eqref{eq:nuus-h_pi} and \eqref{eq:separate-quadratic-surrogate-majorization}, respectively. We empirically observe that the interior condition is generally satisfied if $\alpha$ is appropriately calibrated. However, when a non-negligible portion of $r_i(s)$ reaches $r_{\max}$, the closed-form solution becomes unreliable. We analyze the failures in Appendix~\ref{sec:appendix-additional-experimental-results}. The closed-form solution leads us to a simple evaluation procedure with \method{}: We first sample trajectories to estimate $\kappa$, then evaluate the policy while synthesizing \method{} online at each visited state. Algorithm \ref{alg:evaluate-policy-with-nuus} presents the pseudocode.


\begin{table}[b]
    \fontsize{9pt}{11pt}\selectfont
    \setlength{\tabcolsep}{2pt}
    \centering
    \begin{tabular}{lcccc}
        \toprule
         & PPO & SA-PPO-SGLD & ATLA-PPO & WocaR-PPO \\
        \midrule
        Hopper & 0.15  & 0.25  &  0.0075 & 0.2 \\
        HalfCheetah & 0.06  & 0.2 & 0.02 & 0.075  \\
        Walker2d & 0.025 & 0.075 & 0.05 & 0.15\\
        \bottomrule
    \end{tabular}
    \caption{The coefficient $\alpha$ of the KL divergence regularizer in Eq. \eqref{eq:nuus-surrogate-constraint} for each environment and algorithm during policy evaluation with \method{}.}
    \label{tab:nuus-alpha}
\end{table}

\begin{algorithm}[t]
\caption{Evaluate a policy with \method{}.}
\label{alg:evaluate-policy-with-nuus}
\begin{algorithmic}[1] 
\item[] \hspace{-1.8\algorithmicindent} \textbf{Input}: Policy $\pi$, \method{} hyperparameters $\alpha,\beta$, observation attacker $\nu$. \\
\item[] \hspace{-1.8\algorithmicindent} \textbf{Output}: Trajectories with \method{} synthesized based on each visited state. \\
\STATE Pre-sample trajectories with $\pi$ and estimate $\kappa$ in Eq. \eqref{eq:nuus-solution-kappa-term}.
\FOR{each episode}
\FOR{time step $t = 1\dots T$}
\STATE Observe the current true state $s_t$.
\STATE Calculate $r^*(s_t)$ with Eq. \eqref{eq:nuus-final-solution}.
\STATE Create a \method{} $B_r(s_t)$ in Eq. \eqref{eq:nuus-definition} via $r^*(s_t)$.
\STATE Use attacker $\nu$ to sample a perturbed observed state $\hat{s}_t \sim B_r(s_t)$.
\STATE Predict an action: $a_t \sim \pi(\cdot|\hat{s}_t)$.
\STATE Step forward with $a_t$.
\ENDFOR
\ENDFOR
\end{algorithmic}
\end{algorithm}

\begin{table*}[t]
\centering

\renewcommand{\arraystretch}{0.9}
\fontsize{9pt}{11pt}\selectfont
\setlength{\tabcolsep}{5pt}

\begin{tabular}{cclcccc}
    \toprule[1.2pt]
    \multirow{2}{*}{\textbf{Set}}
    & \multirow{2}{*}{\textbf{Parameter}}
    & \multirow{2}{*}{\textbf{Attacker}}
    & \multicolumn{4}{c}{\textbf{Algorithm}} \\
    \cmidrule(lr){4-7}
    & & & PPO & SA-PPO-SGLD & ATLA-PPO & WocaR-PPO \\
    \midrule

    No Attack & -- & --
    & 421.5 $\pm$ 57.5
    & 349.4 $\pm$ 40.3
    & 345.7 $\pm$ 24.9
    & 331.4 $\pm$ 10.4 \\

    \midrule

    \multirow{16}{*}{$L_\infty$-Norm}
    & \multirow{4}{*}{$\epsilon=0.05$}
    & Random
    & 415.0 $\pm$ 46.7 (6.5)
    & 350.5 $\pm$ 40.7 (-1.1)
    & 342.9 $\pm$ 21.0 (2.8)
    & 331.3 $\pm$ 10.1 (0.1) \\
    &
    & Critic
    & 404.7 $\pm$ 72.6 (16.8)
    & 350.4 $\pm$ 39.3 (-1.0)
    & 344.3 $\pm$ 18.8 (1.4)
    & 331.6 $\pm$ 10.7 (-0.2) \\
    &
    & MAD
    & \textbf{336.7 $\pm$ 47.7 (84.8)}
    & 349.9 $\pm$ 37.5 (-0.5)
    & 339.1 $\pm$ 21.4 (6.6)
    & 328.4 $\pm$ 9.3 (3.0) \\
    &
    & PA-AD
    & 338.3 $\pm$ 67.0 (83.2)
    & \textbf{348.8 $\pm$ 38.6 (0.6)}
    & \textbf{313.1 $\pm$ 38.0 (32.6)}
    & \textbf{324.8 $\pm$ 7.1 (6.6)} \\

    \cmidrule(lr){2-7}

    &
    \multirow{4}{*}{$\epsilon=0.10$}
    & Random
    & 382.9 $\pm$ 44.2 (38.6)
    & 350.6 $\pm$ 37.9 (-1.2)
    & 345.2 $\pm$ 19.7 (0.5)
    & 328.2 $\pm$ 9.0 (3.2) \\
    &
    & Critic
    & 367.3 $\pm$ 98.8 (54.2)
    & 350.5 $\pm$ 41.5 (-1.1)
    & 334.1 $\pm$ 17.5 (11.6)
    & 329.5 $\pm$ 8.1 (1.9) \\
    &
    & MAD
    & 224.6 $\pm$ 58.6 (196.9)
    & 348.2 $\pm$ 40.0 (1.2)
    & 293.0 $\pm$ 42.8 (52.7)
    & \textbf{322.6 $\pm$ 7.0 (8.8)} \\
    &
    & PA-AD
    & \textbf{168.0 $\pm$ 59.8 (253.5)}
    & \textbf{348.0 $\pm$ 42.6 (1.4)}
    & \textbf{243.7 $\pm$ 44.2 (102.0)}
    & 328.4 $\pm$ 8.0 (3.0) \\

    \cmidrule(lr){2-7}

    &
    \multirow{4}{*}{$\epsilon=0.15$}
    & Random
    & 368.3 $\pm$ 36.5 (53.2)
    & 351.2 $\pm$ 38.2 (-1.8)
    & 339.6 $\pm$ 24.5 (6.1)
    & 326.7 $\pm$ 9.8 (4.7) \\
    &
    & Critic
    & 332.2 $\pm$ 92.5 (89.3)
    & 351.1 $\pm$ 39.7 (-1.7)
    & 321.5 $\pm$ 15.7 (24.2)
    & 329.5 $\pm$ 10.1 (1.9) \\
    &
    & MAD
    & 162.1 $\pm$ 55.2 (259.4)
    & 344.6 $\pm$ 37.6 (4.8)
    & 232.8 $\pm$ 39.9 (112.9)
    & 315.3 $\pm$ 8.4 (16.1) \\
    &
    & PA-AD
    & \textbf{100.3 $\pm$ 47.7 (321.2)}
    & \textbf{339.7 $\pm$ 27.3 (9.7)}
    & \textbf{173.8 $\pm$ 53.7 (171.9)}
    & \textbf{300.9 $\pm$ 17.2 (30.5)} \\

    \cmidrule(lr){2-7}

    &
    \multirow{4}{*}{$\epsilon=0.20$}
    & Random
    & 321.8 $\pm$ 38.5 (99.7)
    & 348.5 $\pm$ 40.1 (0.9)
    & 328.3 $\pm$ 26.3 (17.4)
    & 324.3 $\pm$ 8.5 (7.1) \\
    &
    & Critic
    & 280.5 $\pm$ 100.5 (141.0)
    & 347.2 $\pm$ 42.7 (2.2)
    & 295.5 $\pm$ 16.0 (50.2)
    & 324.7 $\pm$ 12.2 (6.7) \\
    &
    & MAD
    & 117.1 $\pm$ 42.6 (304.4)
    & 342.5 $\pm$ 38.2 (6.9)
    & 164.5 $\pm$ 36.9 (181.2)
    & 303.2 $\pm$ 6.8 (28.2) \\
    &
    & PA-AD
    & \textbf{22.3 $\pm$ 33.6 (399.2)}
    & \textbf{337.8 $\pm$ 42.7 (11.6)}
    & \textbf{86.6 $\pm$ 55.7 (259.1)}
    & \textbf{287.0 $\pm$ 12.2 (44.4)} \\

    \midrule

    \multirow{16}{*}{PD-NUUS}
    & \multirow{4}{*}{$\beta=5$}
    & Random
    & 421.2 $\pm$ 60.9 (0.3)
    & 350.0 $\pm$ 39.9 (-0.6)
    & 349.8 $\pm$ 22.7 (-4.1)
    & 330.1 $\pm$ 11.9 (1.3) \\
    &
    & Critic
    & 418.1 $\pm$ 56.6 (3.4)
    & 350.4 $\pm$ 41.3 (-1.0)
    & 350.2 $\pm$ 20.1 (-4.5)
    & 332.1 $\pm$ 9.4 (-0.7) \\
    &
    & MAD
    & \textbf{409.2 $\pm$ 65.3 (12.3)}
    & 349.6 $\pm$ 41.4 (-0.2)
    & 349.4 $\pm$ 20.0 (-3.7)
    & \textbf{329.6 $\pm$ 9.2 (1.8)} \\
    &
    & PA-AD
    & 418.2 $\pm$ 60.8 (3.3)
    & \textbf{346.2 $\pm$ 41.2 (3.2)}
    & \textbf{347.9 $\pm$ 21.7 (-2.2)}
    & 329.9 $\pm$ 8.9 (1.5) \\
    
    \cmidrule(lr){2-7}
    
    &
    \multirow{4}{*}{$\beta=10$}
    & Random
    & \textbf{417.6 $\pm$ 58.1 (3.9)}
    & 348.3 $\pm$ 41.2 (1.1)
    & 350.1 $\pm$ 19.2 (-4.4)
    & 330.4 $\pm$ 9.6 (1.0) \\
    &
    & Critic
    & 417.9 $\pm$ 59.9 (3.6)
    & 351.5 $\pm$ 40.4 (-2.1)
    & 351.4 $\pm$ 18.3 (-5.7)
    & 333.4 $\pm$ 10.3 (-2.0) \\
    &
    & MAD
    & 422.4 $\pm$ 60.0 (-0.9)
    & \textbf{348.1 $\pm$ 38.8 (1.3)}
    & 346.3 $\pm$ 23.8 (-0.6)
    & 329.2 $\pm$ 11.4 (2.2) \\
    &
    & PA-AD
    & 424.4 $\pm$ 58.1 (-2.9)
    & 348.6 $\pm$ 38.8 (0.8)
    & \textbf{344.2 $\pm$ 21.2 (1.5)}
    & \textbf{325.6 $\pm$ 8.6 (5.8)} \\
    
    \cmidrule(lr){2-7}
    
    &
    \multirow{4}{*}{$\beta=20$}
    & Random
    & 409.3 $\pm$ 59.1 (12.2)
    & 347.7 $\pm$ 39.3 (1.7)
    & 350.3 $\pm$ 21.2 (-4.6)
    & 329.8 $\pm$ 10.2 (1.6) \\
    &
    & Critic
    & 414.8 $\pm$ 65.6 (6.7)
    & 350.5 $\pm$ 40.4 (-1.1)
    & 346.1 $\pm$ 20.4 (-0.4)
    & 331.4 $\pm$ 11.2 (0.0) \\
    &
    & MAD
    & 413.9 $\pm$ 68.6 (7.6)
    & 346.5 $\pm$ 36.8 (2.9)
    & 344.1 $\pm$ 24.5 (1.6)
    & 326.8 $\pm$ 12.2 (4.6) \\
    &
    & PA-AD
    & \textbf{404.3 $\pm$ 54.5 (17.2)}
    & \textbf{345.2 $\pm$ 38.7 (4.2)}
    & \textbf{332.3 $\pm$ 19.0 (13.4)}
    & \textbf{311.5 $\pm$ 13.6 (19.9)} \\
    
    \cmidrule(lr){2-7}
    
    &
    \multirow{4}{*}{$\beta=40$}
    & Random
    & 416.7 $\pm$ 59.0 (4.8)
    & 347.1 $\pm$ 38.1 (2.3)
    & 349.6 $\pm$ 20.7 (-3.9)
    & 321.3 $\pm$ 12.8 (10.1) \\
    &
    & Critic
    & 413.6 $\pm$ 60.6 (7.9)
    & 340.4 $\pm$ 36.2 (9.0)
    & 335.2 $\pm$ 23.4 (10.5)
    & 324.0 $\pm$ 9.1 (7.4) \\
    &
    & MAD
    & 405.4 $\pm$ 61.6 (16.1)
    & 330.2 $\pm$ 29.4 (19.2)
    & 335.8 $\pm$ 24.6 (9.9)
    & 310.2 $\pm$ 11.9 (21.2) \\
    &
    & PA-AD
    & \textbf{387.3 $\pm$ 68.3 (34.2)}
    & \textbf{327.0 $\pm$ 44.3 (22.4)}
    & \textbf{319.6 $\pm$ 19.1 (26.1)}
    & \textbf{287.0 $\pm$ 24.8 (44.4)} \\

    \bottomrule[1.2pt]
\end{tabular}
\caption{Comparison of $L_\infty$-norm uncertainty sets and \method{} on HalfCheetah. Each entry reports the discounted return, with degradation in parentheses. The largest degradation across attackers is in \textbf{bold}. The episode rewards of clean policies are 5275.1 $\pm$ 1586.0 (PPO), 4320.3 $\pm$ 611.3 (SA-PPO-SGLD), 4267.1 $\pm$ 854.5 (ATLA-PPO) and 3952.2 $\pm$ 566.3 (WocaR-PPO).}
\label{tab:halfcheetah-discounted-return-degradation-in-main}
\end{table*}

\section{Experimental Results}

In this section, we aim to answer the following questions:
\begin{itemize}
    \item Does \method{} produce different perturbation scales for different policy algorithms?
    \item Is $\beta$ correlated with the empirical worst-case performance degradation, and how does this relationship compare with that of an $L_\infty$-norm uncertainty set?
    \item Does addressing the discrepancy between $d^{\pi_\nu}$ and $ d^\pi$ with non-zero $\alpha$ improve the quality of \method{}?
\end{itemize}

\subsection{Experiment Setup}

\begin{table*}[t]
    \centering
    {

    \renewcommand{\arraystretch}{1.2}
    \fontsize{9pt}{11pt}\selectfont
    \setlength{\tabcolsep}{5pt}
    \begin{tabular}{llcccccccc}
        \toprule[1.2pt]
        \multirow{2}{*}{\textbf{Environment}} & \multirow{2}{*}{\textbf{Algorithm}} & \multicolumn{4}{c}{$L_\infty$-Norm} & \multicolumn{4}{c}{\method{}} \\
        \cmidrule(lr){3-6} \cmidrule(lr){7-10}
         & & $\epsilon=0.05$ & $\epsilon=0.10$ & $\epsilon=0.15$ & $\epsilon=0.20$ & $\beta=5$ & $\beta=10$ & $\beta=20$ & $\beta=40$ \\ \hline

         \multirow{4}{*}{Hopper} & PPO & 52.0 & 162.3 & 168.9 & 198.5 & 1.7 & 10.0 & 2.7 & 21.0 \\
          & SA-PPO-SGLD & 0.6 & 0.6 & 0.6 & 28.4 & 3.9 & 5.8 & 14.6 & 30.8 \\
          & ATLA-PPO & 17.3 & 78.7 & 103.0 & 136.8 & 1.0 & 2.5 & 24.5 & 127.7 \\
          & WocaR-PPO & 1.2 & 10.6 & 25.7 & 151.7 & 2.7 & 2.4 & 9.3 & 21.7 \\ \hline

          \multirow{4}{*}{HalfCheetah} & PPO & 84.8 & 253.5 & 321.2 & 399.2 & 12.3 & 3.9 & 17.2 & 34.2 \\
          & SA-PPO-SGLD & 0.6 & 1.4 & 9.7 & 11.6 & 3.2 & 1.3 & 4.2 & 22.4 \\
          & ATLA-PPO & 32.6 & 102.0 & 171.9 & 259.1 & -2.2 & 1.5 & 13.4 & 26.1 \\
          & WocaR-PPO & 6.6 & 8.8 & 30.5 & 44.4 & 1.8 & 5.8 & 19.9 & 44.4 \\ \hline

          \multirow{4}{*}{Walker2d} & PPO & 50.3 & 139.0 & 162.3 & 205.3 & 2.8 & 7.7 & 60.5 & 105.6 \\
          & SA-PPO-SGLD & 1.0 & 23.0 & 50.3 & 128.9 & 16.0 & 14.9 & 15.4 & 128.3 \\
          & ATLA-PPO & 49.0 & 125.7 & 165.1 & 191.5 & 0.1 & 3.2 & 7.9 & 12.3 \\
          & WocaR-PPO & 2.9 & 42.3 & 135.2 & 139.0 & 5.6 & 1.7 & 16.7 & 50.0 \\
          \bottomrule[1.2pt]
    \end{tabular}
    }
    \caption{Maximum empirical degradation in discounted return, observed among all state attackers for each environment, algorithm and uncertainty set parameter.}
    \label{tab:worst-case-degradation-and-correlation}
\end{table*}

\subsubsection{Algorithms and Environments.} We choose PPO \cite{schulman2017proximal}, SA-PPO-SGLD \cite{zhang2020robust}, ATLA-PPO \cite{zhang2021robust}, and WocaR-PPO \cite{liang2022efficient} for policy training. Except for PPO, all algorithms incorporate dedicated mechanisms for improving robustness against state perturbations. We conduct experiments on Hopper, HalfCheetah, and Walker2d environments from MuJoCo \citep{todorov2012mujoco}. For each environment and algorithm, we train policies using four random seeds and report the average results. Since synthesizing \method{} requires estimates of the advantage function $A^\pi(s,a)$, we train an auxiliary advantage network alongside each policy without changing the underlying policy update rule. The discount factor is fixed at $\gamma=0.99$ throughout the experiments.

\subsubsection{State Attackers.} We consider four attackers that perturb the observed states within a given uncertainty set:
\begin{itemize}
    \item Random: Uniformly samples from the uncertainty set.
    \item Critic \cite{zhang2020robust}: Uses projected gradient descent (PGD) to search for an adversarial state that minimizes the state-value function.
    \item Maximal Action Difference (MAD) \cite{zhang2020robust}: A critic-free attack that searches for an adversarial state $\hat{s}$ by maximizing the KL divergence $D_{\text{KL}}(\pi(\cdot|s)||\pi(\cdot|\hat{s}))$.
    \item PA-AD \cite{sun2022who}: Formulates state perturbations as a Policy Adversary MDP (PAMDP), where a director learns to select an adversarial action, and an actor finds a feasible state perturbation that induces the desired policy change. An optimal policy of the PAMDP theoretically induces a state adversary that minimizes the clean policy's expected discounted return.
\end{itemize}
According to \citet{zhang2020robust,liang2022efficient}, Random and Critic are weaker than PA-AD and MAD. All four attacks were originally formulated for $L_\infty$-norm uncertainty sets \citep{zhang2020robust,zhang2021robust,sun2022who,liang2022efficient}. To adapt them to \method{}, we replace the budget $\epsilon$ with $r_i(s)$ for every state dimension $i$. 


\subsubsection{Implementation Details.} For \method{}, Table \ref{tab:nuus-alpha} reports the value of $\alpha$ for each environment and algorithm, and we clip the solution $r(s)$ to $[0.001, 2.5]$. We parameterize the policy $\pi$ with a Gaussian distribution $\mathcal{N}(\mu_{\theta}(s),\Sigma)$, where $\mu_{\theta}$ is a multilayer perceptron for predicting the mean and $\Sigma=\text{diag}(\sigma_1^2,\dots,\sigma^2_n)$ is a diagonal covariance matrix. We exploit this Gaussian policy structure to compute the state Fisher information matrix and approximate the Hessian term in $H_{\pi,\alpha}(s)$ in Eq. \eqref{eq:nuus-reformulation-H-matrix}, using a Gauss-Newton approximation. More implementation details are provided in Appendix~\ref{sec:appendix-additional-implementation-details}.

\subsection{Empirical Results}

\subsubsection{$L_\infty$-Norm vs. \method{}.} We report the empirical $\kappa$ values for synthesizing PD-NUUS in Appendix~\ref{sec:appendix-additional-experimental-results}. Table \ref{tab:halfcheetah-discounted-return-degradation-in-main} compares the empirical degradation in discounted return on HalfCheetah under $L_\infty$-norm uncertainty sets with $\epsilon \in \{0.05, 0.1, 0.15, 0.2\}$ and \method{} with $\beta \in \{ 5,10,20,40 \}$. Results for Hopper and Walker2d are provided in Appendix~\ref{sec:appendix-additional-experimental-results}. We use the maximum degradation observed among all attackers as an empirical approximation of the worst-case degradation. As expected, increasing $\epsilon$ enlarges the $L_\infty$-norm uncertainty set and generally admits stronger perturbations. However, the relationship between $\epsilon$ and return degradation is highly algorithm-dependent: SA-PPO-SGLD and WocaR-PPO experience substantially smaller degradation than PPO and ATLA-PPO under the same $\epsilon$. In contrast, the tolerance $\beta$ of \method{} is specified in units of discounted return. The maximum degradation across attackers is better aligned with $\beta$ than with $\epsilon$, and generally remains below or moderately above the tolerance $\beta$.

\subsubsection{Tolerance Calibration.} 
Table \ref{tab:worst-case-degradation-and-correlation} summarizes the empirical maximum degradation in discounted return among all attackers. Although several failure cases occur when $\beta=40$, \method{} generally provides more reliable control of empirical degradation for different algorithms. We next quantitatively evaluate whether the tolerance $\beta$ controls the maximum degradation in Table \ref{tab:worst-case-degradation-and-correlation}. Requiring the maximum degradation to stay strictly below $\beta$ as in Eq. \eqref{eq:ideal-performance-degradation-constraint} is too restrictive, because the surrogate constraint, local approximation errors, SQS majorization and finite-horizon trajectories can introduce discrepancies between the predefined and observed degradations. Therefore, we define the \emph{relaxed-tolerance frequency}
\begin{equation}
    C(\tau) = \frac{1}{N} \sum^N_{j=1} \mathbb{I} [\Delta\widehat{J}_{\max,j} \leq (1+\tau)\beta], \ \tau\geq0
\end{equation}
where $N=48$ is the total number of evaluations with \method{}, $\Delta\widehat{J}_{\max,j}$ is the empirical maximum degradation in setting $j$, and $\mathbb{I}[X]=1$ if $X$ is true and 0 otherwise. Thus, $C(\tau)$ denotes the fraction of settings in which the empirical maximum degradation exceeds the tolerance $\beta$ by no more than a relative margin $\tau$. Table \ref{tab:relaxed-tolerance} indicates that $\beta$ provides informative control over the scale of empirical degradation.

\begin{table}[!htbp]
    \centering
     \renewcommand{\arraystretch}{1.3}
    \fontsize{10pt}{11pt}\selectfont
    \setlength{\tabcolsep}{5pt}
    \begin{tabular}{ccccc} \toprule[1.2pt]
        $\tau$ & 0 & 0.1 & 0.2 & 0.25 \\ \midrule
        $C(\tau)$ & 77.08\% & 77.08\% & 81.25\% & 85.42\% \\ \bottomrule[1.2pt]
    \end{tabular}
    \caption{$C(\tau)$ values calculated across all \method{} tests.}
    \label{tab:relaxed-tolerance}
\end{table}



\begin{figure*}[t]
    \centering
    \includegraphics[scale=0.44]{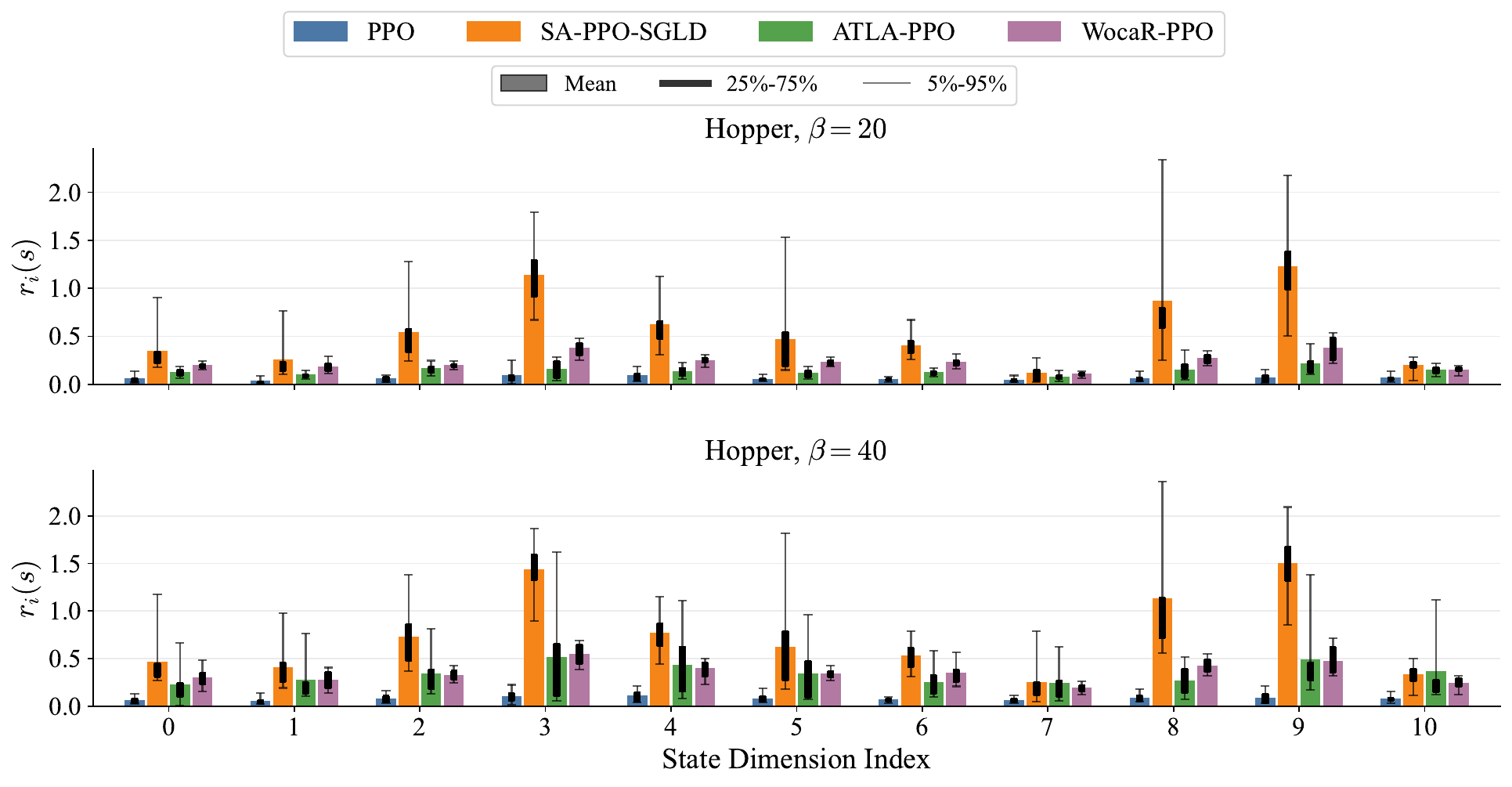}
    \caption{We report the mean and confidence range of the calculated budget function $r_i(s)$ for each state dimension across four algorithms and $\beta \in \{20,40\}$ on Hopper under PA-AD attack. The statistics are computed over 200 sampled trajectories.}
    \label{fig:budgets}
\end{figure*}

\subsubsection{Ablation Study of the KL Divergence Regularizer.} We conduct an ablation study to evaluate the role of the KL divergence regularizer in Eq. \eqref{eq:nuus-surrogate-constraint}. Omitting the regularizer induces a solution that completely ignores the occupancy shift for any $\beta$ value (assuming $d^{\pi_\nu}\approx d^\pi$). Table \ref{tab:ablation-study} shows that the policy performance degrades substantially without the KL divergence regularizer for both $\beta$ values and all algorithms.

\begin{table}[t]
    \centering
    \fontsize{9pt}{11pt}\selectfont
    \setlength{\tabcolsep}{3.5pt}
    \begin{tabular}{llcccc}
        \toprule[1.2pt]
        \multirow{2}{*}{\textbf{Environment}}
        & \multirow{2}{*}{\textbf{Algorithm}}
        & \multicolumn{2}{c}{$\beta=20$}
        & \multicolumn{2}{c}{$\beta=40$} \\
        \cmidrule(lr){3-4} \cmidrule(lr){5-6}
        & & w/ & w/o & w/ & w/o \\
        \midrule

        \multirow{4}{*}{Hopper}
        & PPO         & 2.7  & 228.7 & 21.0  & 241.4 \\
        & SA-PPO-SGLD & 14.6 & 55.0  & 30.8  & 72.4  \\
        & ATLA-PPO    & 24.5 & 128.1 & 127.7 & 176.7 \\
        & WocaR-PPO   & 9.3  & 134.5 & 21.7  & 183.2 \\
        \midrule

        \multirow{4}{*}{HalfCheetah}
        & PPO         & 17.2 & 285.3 & 34.2 & 341.7 \\
        & SA-PPO-SGLD & 4.2  & 112.8 & 22.4 & 151.5 \\
        & ATLA-PPO    & 13.4 & 245.1 & 26.1 & 284.3 \\
        & WocaR-PPO   & 19.9 & 129.4 & 44.4 & 157.5 \\
        \midrule

        \multirow{4}{*}{Walker2d}
        & PPO         & 60.5 & 256.0 & 105.6 & 273.4 \\
        & SA-PPO-SGLD & 15.4 & 252.9 & 128.3 & 251.9 \\
        & ATLA-PPO    & 7.9  & 211.5 & 12.3  & 232.5 \\
        & WocaR-PPO   & 16.7 & 218.7 & 50.0  & 203.4 \\

        \bottomrule[1.2pt]
    \end{tabular}

    
    \caption{Empirical degradation $J(\pi)-J(\pi_\nu)$ in discounted total returns using \method{} with (w/) and without (w/o) the KL divergence regularizer. With regularization, we report the maximum empirical degradation across four attackers; without regularization ($\alpha=0$), we report degradation under PA-AD. Even under this conservative comparison, degradation without regularization substantially exceeds $\beta$.}

    \label{tab:ablation-study}
\end{table}

\subsubsection{\method{} Visualization Results.} In Figure \ref{fig:budgets}, we visualize the synthesized \method{} budgets $r_i(s)$ for four algorithms under PA-AD attack. As $\beta$ increases, the budgets generally increase for all algorithms. SA-PPO-SGLD admits larger perturbations than other algorithms nearly for all dimensions. Under PA-AD attack, PD-NUUS at $\beta=40$ reduces episode rewards from 3471.7 $\pm$ 391.7 to 2489.8 $\pm$ 1287.3 for PPO, from 3203.0 $\pm$ 378.5 to 1198.2 $\pm$ 1085.3 for SA-PPO-SGLD, from 3227.5 $\pm$ 634.9 to 924.3 $\pm$ 1323.3 for ATLA-PPO, and from 3358.3 $\pm$ 452.8 to 1663.1 $\pm$ 1081.5 for WocaR-PPO compared with no attack.

\section{Limitations and Conclusion}
We discuss several limitations of our work here. First, \method{} selects the degradation in expected discounted return as the criterion. However, the discount factor significantly underweight rewards at later time steps. To address this issue, we may extend \method{} to average-return scenarios. In addition, while the surrogate constraint makes the problem tractable, it introduces additional approximations. Finally, in this work we only use \method{} for policy evaluation, and the effects of \method{} on policy training remain unknown. Meanwhile for training, the pre-sampling phase at early training stages may result in unreliable \method{}. Future work should address these concerns for robust policy training.

In summary, we introduce \method{}, an interpretable uncertainty set formulation for observed state perturbations. By formulating its synthesis as a constrained optimization problem, we derive an efficient closed-form solution for analyzing policy robustness. In empirical results, \method{} provides anisotropic uncertainty sets with generally constrained performance degradation.

\appendix


\bibliography{aaai2027}


\newpage

\section{Proofs}

\subsection{Proof of Lemma~\ref{lemma:occupancy-shift-on-nuus}}
\label{proof:occupancy-shift-on-nuus}

\paragraph{Lemma 1.} {\itshape
Assume that a transition kernel is $L_P$-Lipschitz continuous w.r.t the action: $||P(\cdot|s,a_1)-P(\cdot|s,a_2)||_1\leq L_P ||a_1-a_2||_1$ for all $s\in \mathcal{S}$ and for all $a_1,a_2 \in \mathcal{A}$, and a stochastic policy $\pi$ is $L_\pi$-Lipschitz continuous from states to action distributions: $W_1(\pi(\cdot|s_1),\pi(\cdot|s_2)) \leq L_\pi ||s_1-s_2||_\infty$ for all $s_1,s_2\in \mathcal{S}$ where $W_1$ is the 1-Wasserstein distance induced by the $L_1$-norm on the action space $\mathcal{A}$. Then for any state adversary $\nu \in \mathcal{V}_r$,
\begin{equation}
    D_{\text{TV}}(d^{\pi},d^{\pi_\nu}) \leq \frac{\gamma L_PL_\pi r_{\max}}{2(1-\gamma)}.
\end{equation}
}

\begin{proof}
The proof is motivated by \citet{achiam2017constrained,mao2026adaptive}. Define
\begin{gather}
P^\pi(s'| s) \triangleq \int_{a\in \mathcal{A}} P(s'|s,a)\pi(a|s) \ da.
\end{gather}
The 1-Wasserstein induced by $L_1$-norm on the action space $\mathcal{A}$ is expressed as
\begin{equation}
    W_1(\xi_1,\xi_2) \triangleq \inf_{\zeta \in \Gamma(\xi_1,\xi_2)} \int ||a-a'||_1 \ \zeta(da,da'),
\end{equation}
where $\Gamma(\xi_1,\xi_2)$ denotes the set of couplings whose marginals are $\xi_1$ and $\xi_2$. Then, for any coupling $\zeta_s \in \Gamma(\pi(\cdot|s),\pi_\nu(\cdot|s))$ we upper bound the 1-norm distance
\begin{align}
&\quad ||P^\pi(\cdot\mid s)-
P^{\pi_\nu}(\cdot\mid s)||_1 \\
&=  \left\| \int_{A} P(s'|s,a)\pi(da|s) - \int_{A} P(s'|s,a')\pi_\nu(da'|s) \right\|_1 \\
&=  \left\| \int_{\mathcal{A}\times \mathcal{A}} [P(\cdot|s,a) - P(\cdot|s,a')] \ \zeta_s(da,da') \right\|_1  \\
&\leq  \int_{\mathcal{A}\times \mathcal{A}} ||P(\cdot|s,a) - P(\cdot|s,a')||_1 \ \zeta_s(da,da') \\
&\leq L_P \int_{\mathcal{A}\times \mathcal{A}} ||a-a
'||_1 \ \zeta_s(da,da'),
\end{align} 
which holds for every coupling $\zeta_s$. Besides, for $\nu \in \mathcal{V}_r$ we have $||s - \nu(s)||_\infty \leq ||r(s)||_\infty \leq r_{\max}$ for $\forall s\in \mathcal{S}$. Taking the infimum over all couplings yields
\begin{align}
D_{\text{TV}}(P^\pi(\cdot\mid s),
P^{\pi_\nu}(\cdot\mid s)) 
&\leq L_P W_1(\pi(\cdot|s),\pi_\nu(\cdot|s)) \\
&\leq L_PL_\pi ||s-\nu(s)||_\infty \\
&\leq L_PL_\pi r_{\max} . \label{eq:bound-of-Ppi-Ppinu}
\end{align}
The occupancy $d^\pi$ is also expressed as:
\begin{equation}
    d^\pi (s) = (1-\gamma)\rho + \gamma
\int_{s' \in \mathcal{S}}
P^{\pi}(s|s')d^{\pi}(s') \ ds.
\end{equation}
Then
\begin{equation}
\begin{aligned}
&\quad d^\pi(s)-d^{\pi_\nu}(s) \nonumber \\
&= \gamma \int_{s'} \Big[ P^\pi(s| s')d^\pi(s') - P^{\pi_\nu}(s| s')d^{\pi_\nu}(s') \Big] \ ds' \\
&= \gamma \int_{s'}  P^\pi(s| s')[d^\pi(s') -d^{\pi_\nu}(s')] \ ds' \nonumber \\
&\quad \ \ + \gamma \int_{s'} [P^{\pi}(s| s') - P^{\pi_\nu}(s| s')]d^{\pi_\nu}(s')  \ ds'.
\end{aligned}
\end{equation}
Let $\Delta (s) \triangleq d^\pi(s)-d^{\pi_\nu}(s)$ then use the triangle inequality
\begin{align}
&\quad \ ||\Delta||_1 \nonumber \\
&:= \int_{s} |\Delta(s)| \ ds \\
&\leq \gamma \int_{s} \int_{s'}  P^\pi(s| s')|d^\pi(s') -d^{\pi_\nu}(s')| \ ds' ds \nonumber \\
&\quad \ \ + \gamma \int_{s} \int_{s'} |P^{\pi}(s| s') - P^{\pi_\nu}(s| s')|d^{\pi_\nu}(s')  \ ds'ds \\
&= \gamma \int_{s'} |\Delta(s')| \Big[ \int_{s}  P^\pi(s| s') \ ds \Big] ds' \nonumber \\
&\quad \ \ + \gamma  \int_{s'}d^{\pi_\nu}(s') \Big[ \int_{s}|P^{\pi}(s| s') - P^{\pi_\nu}(s| s')|  \ ds \Big] ds' \\
&\leq \gamma ||\Delta||_1 + \gamma  \int_{s'}d^{\pi_\nu}(s') ||P^{\pi}(\cdot| s') - P^{\pi_\nu}(\cdot| s')||_1 ds' \\
&\leq \gamma ||\Delta||_1 + \gamma \sup_{s' \in \mathcal{S}} ||P^{\pi}(\cdot| s') - P^{\pi_\nu}(\cdot| s')||_1.
\end{align}
This indicates
\begin{equation}
    ||\Delta||_1 \leq \frac{\gamma}{1-\gamma} \sup_{s' \in \mathcal{S}} ||P^{\pi}(\cdot| s') - P^{\pi_\nu}(\cdot| s')||_1.
\end{equation}
Since $D_{\text{TV}}(d^\pi,d^{\pi_\nu}) = \frac{1}{2}||\Delta||_1$, from Eq. \eqref{eq:bound-of-Ppi-Ppinu} we get
\begin{align}
D_{\text{TV}}(d^\pi,d^{\pi_\nu}) 
&\leq \frac{\gamma}{2(1-\gamma)} \sup_{s' \in \mathcal{S}} ||P^{\pi}(\cdot| s') - P^{\pi_\nu}(\cdot| s')||_1 \\
&\leq \frac{\gamma L_P L_\pi r_{\max}}{2(1-\gamma)}.
\end{align}
This completes the proof.

\end{proof}

\subsection{Proof of Lemma~\ref{lemma:degradation-upperbound-on-nuus}}

\paragraph{Lemma 2.} {\itshape
Consider a fixed policy $\pi$, $\gamma\in(0,1)$, and $\alpha,\beta>0$. After using the surrogate constraint in Eq. \eqref{eq:nuus-surrogate-constraint}, the supremum degradation of discounted return satisfies the following upper bound
\begin{equation}
\sup_{\nu\in\mathcal V_{r}}
\bigl[J(\pi)-J(\pi_\nu)\bigr]-\beta
\le
\frac{\gamma^2(\max_{s,a} |A^\pi(s,a)|)^2}{2\alpha(1-\gamma)^3}.
\end{equation}
}

\label{proof:degradation-upperbound-on-nuus}
\begin{proof}
Fix an arbitrary adversary $\nu\in\mathcal V_{r}$ and denote
\begin{gather}
\ell_\nu(s) = -\mathbb E_{a\sim\pi_\nu(\cdot\mid s)} [A^\pi(s,a)],\\
K_\nu = \mathbb E_{s\sim d^\pi} \left[ D_{\mathrm{KL}} \bigl(\pi_\nu(\cdot\mid s)|\pi(\cdot\mid s)\bigr) \right],\\
S_\alpha(\nu) = \mathbb E_{s\sim d^\pi}[\ell_\nu(s)] +\alpha K_\nu.
\end{gather}
Surrogate feasibility implies
\begin{equation}
S_\alpha(\nu)\le(1-\gamma)\beta, \ \  \forall \nu \in \mathcal{V}_r.
\end{equation}
By the performance difference lemma,
\begin{equation}
J(\pi)-J(\pi_\nu)= \frac{1}{1-\gamma} \mathbb E_{s\sim d^{\pi_\nu}}[\ell_\nu(s)].
\end{equation}
Denote $G=\max_{s,a}|A^\pi(s,a)|$, for every $\nu\in\mathcal V_r$,
\begin{equation}
\begin{aligned}
\max_s \left|
\mathbb E_{a\sim\pi_\nu(\cdot\mid s)}[A^\pi(s,a)]
\right|
&\leq
\max_s \mathbb E_{a\sim\pi_\nu(\cdot\mid s)}
\left[|A^\pi(s,a)|\right]\\
&\leq \max_{s,a}|A^\pi(s,a)| = G.
\end{aligned}
\end{equation}
Applying Corollary 1 and 3 of \cite{achiam2017constrained} with $\pi'=\pi_\nu$, together with $G$, gives
\begin{equation}
\mathbb E_{s\sim d^{\pi_\nu}}[\ell_\nu(s)] \leq \mathbb E_{s\sim d^\pi}[\ell_\nu(s)] +C\sqrt{K_\nu},
\end{equation}
where
\begin{equation}
C=\frac{\sqrt{2}\gamma G}{1-\gamma}.
\end{equation}
For $\alpha>0$, Young's inequality yields
\begin{equation}
C\sqrt{K_\nu}
\le
\alpha K_\nu+\frac{C^2}{4\alpha}.
\end{equation}
Equivalently, this follows by expanding $(\sqrt{\alpha K_\nu}-C/(2\sqrt{\alpha}))^2\ge0$. Then
\begin{equation}
\begin{aligned}
(1-\gamma)[J(\pi)-J(\pi_\nu)] 
&\leq \mathbb E_{s\sim d^\pi}[\ell_\nu(s)] +C\sqrt{K_\nu} \\
&= \mathbb E_{s\sim d^\pi}[\ell_\nu(s)]  + \alpha K_\nu +\frac{C^2}{4\alpha} \\
&\leq S_\alpha(\nu)+\frac{C^2}{4\alpha}\\
&\le
(1-\gamma)\beta+\frac{C^2}{4\alpha}.
\end{aligned}
\end{equation}
Dividing by $1-\gamma$ and substituting the definition of $C$ gives
\begin{equation}
J(\pi)-J(\pi_\nu)-\beta
\le
\frac{\gamma^2G^2}{2\alpha(1-\gamma)^3}.
\end{equation}
Since $\nu$ was arbitrary and $G$ is independent of the adversary class $\mathcal{V}_r$, taking the supremum over $\nu\in\mathcal V_{r}$ proves the lemma.
\end{proof}

\section{Derivation of a Closed-Form Solution}
\label{sec:appendix-derive-solution}

\subsection{Reformulate the Surrogate Constraint}
\label{sec:appendix-reformulate-surrogate-constraint}
The original constrained optimization problem is intractable because it relies on $d^{\pi_\nu}$ which requires the knowledge of both the transition dynamics and the perturbed policy $\pi_\nu$, while the exact $\pi_\nu$ is unknown but only known to be constrained by a \method{} $B_r(s)$ and an admissible set $\mathcal{V}_r$ of state adversary
\begin{gather}
\mathcal{V}_r \triangleq \{\nu:\mathcal{S}\rightarrow\mathcal{S} | \ \nu(s)\in B_r(s), \ \forall s\in\mathcal{S} \}, \\
B_r(s) \triangleq \left\{ \hat{s}\in\mathcal{S}| \ \ |s_i-\hat{s}_i| \leq r_i(s), \ i=1,\dots,m \right\}.
    \label{eq:nuus-definition}
\end{gather}
We propose a surrogate constraint
\begin{equation}
    \sup_{\nu \in \mathcal{V}_r} -\mathbb{E}_{s\sim d^{\pi}} \left[\mathbb{E}_{a\sim \pi_\nu} [A^\pi(s,a)]- \alpha D_{\text{KL}} (\pi_\nu||\pi)[s] \right]
\end{equation}
to make the problem tractable. This constraint is still dependent on $\nu$, and we express it as a function of $r(s)$. 

Let $\delta(s) = \nu(s) - s$, then the perturbed policy $\pi_\nu$ is approximated by a second-order Taylor expansion
\begin{align}
& \ \quad \pi_\nu(a|s) \\
&= \pi(a|s) + \nabla_s \pi(a|s)^\top \delta(s) \nonumber \\ 
&\quad \quad + \frac{1}{2} \delta(s)^\top \nabla_s^2\pi(a|s) \delta(s) + o(||\delta||_2^2) \\
&\approx  \pi(a|s) \Big[ 1 + \nabla_s \log \pi(a|s)^\top \delta(s)  \nonumber \\
&\quad \quad + \frac{1}{2} \delta(s)^\top \Big(  \nabla_s \log\pi(a|s) \nabla_s \log\pi(a|s)^\top \nonumber \\
& \qquad \qquad + \nabla_s^2 \log\pi(a|s) \Big) \delta(s) \Big].
\end{align}
The KL divergence is approximated by using the state Fisher information matrix $F_\pi(s)$
\begin{gather}
D_{\text{KL}}(\pi_\nu || \pi)[s] \approx \frac{1}{2} \delta(s)^\top F_\pi(s)  \delta(s), \\
F_\pi(s) = \mathbb{E}_{a\sim \pi}[\nabla_s \log \pi(a|s) \nabla_s \log \pi(a|s)^\top].
\end{gather}
Since $\mathbb{E}_{a\sim \pi}[A^\pi(s,a)]=0$, we get that
\begin{align}
& \quad \ \mathbb{E}_{a\sim \pi_\nu} [A^\pi(s,a)]- \alpha D_{\text{KL}} (\pi_\nu||\pi)[s] \nonumber \\
&= \mathbb{E}_{a\sim \pi} \Big[ \frac{\pi_\nu(a|s)}{\pi(a|s)} A^\pi(s,a) \Big] - \alpha D_{\text{KL}} (\pi_\nu||\pi)[s] \\
&\approx \mathbb{E}_{a\sim \pi} \Big[  A^\pi(s,a) \nabla_s \log \pi(a|s)^\top \delta(s)  \nonumber \\
&\quad \quad + \frac{A^\pi(s,a)}{2} \delta(s)^\top \Big(  \nabla_s \log\pi(a|s) \nabla_s \log\pi(a|s)^\top \nonumber \\
& \quad \quad + \nabla_s^2 \log\pi(a|s) \Big) \delta(s)  \Big] -  \frac{\alpha}{2} \delta(s)^\top F_\pi(s)  \delta(s)  \\
&= -\frac{1}{2}\delta(s)^\top H_{\pi,\alpha}(s) \delta(s) + g_\pi(s)^\top \delta(s),
\end{align}
where
\begin{align}
& H_{\pi,\alpha}(s) = \alpha F_\pi(s) -\mathbb{E}_{a\sim \pi} \Big[ A^\pi(s,a) \Big( \nonumber \\
& \qquad \nabla_s \log\pi(a|s) \nabla_s \log\pi(a|s)^\top + \nabla_s^2 \log\pi(a|s) \Big) \Big], \label{eq:nuus-reformulation-H-matrix}
\end{align}
and
\begin{equation}
    g_\pi(s) = \mathbb{E}_{a\sim \pi}[A^\pi(s,a) \nabla_s \log \pi(a|s)] \label{eq:nuus-g_pi}.
\end{equation}
Then the surrogate constraint is equivalent to
\begin{equation}
    \mathbb{E}_{s\sim d^{\pi}}\left[ \sup_{\delta(s)} \left\{ \frac{1}{2}\delta(s)^\top H_{\pi,\alpha}(s) \delta(s) - g_\pi(s)^\top \delta(s) \right\} \right]
\end{equation}
where $|\delta_i(s)| \leq r_i(s)$ holds for each state dimension $i$. However, $H_{\pi,\alpha}(s)$ is not necessarily positive semidefinite, and if this happens then the problem generally does not have a closed-form solution. To derive a closed-form solution, we construct a positive-definite matrix $M_{\pi,\alpha}$ satisfying both $M_{\pi,\alpha}(s) \succ 0$ and $M_{\pi,\alpha}(s) \succeq H_{\pi,\alpha}(s)$.  This indicates that $\delta(s)^\top M_{\pi,\alpha}(s)\delta(s) \geq \delta(s)^\top H_{\pi,\alpha}(s)\delta(s)$ holds for every $\delta(s)$. We implement $M_{\pi,\alpha}(s)$ using SQS majorization as in Section 4.2. Then we obtain a convex constraint by:
\begin{align} 
& \quad \  \sup_{\delta(s)} \left\{ \frac{1}{2}\delta(s)^\top H_{\pi,\alpha}(s) \delta(s) - g_\pi(s)^\top \delta(s) \right\} \\ 
& \leq  \sup_{\delta(s)} \left\{ \frac{1}{2}\delta(s)^\top M_{\pi,\alpha}(s) \delta(s) - g_\pi(s)^\top \delta(s) \right\}  \\
&= \frac{1}{2}r(s)^\top M_{\pi,\alpha}(s) r(s) + |g_\pi(s)|^\top r(s).
\end{align}
Finally, the problem of \method{} is reformulated as a tractable constrained problem:
\begin{align}
\max_r & \quad  \mathbb{E}_{s\sim d^\pi} [ \textbf{1}^\top r(s)], \\ 
\text{s.t.} & \quad \mathbb{E}_{s\sim d^{\pi}}\Big[  \frac{1}{2}r(s)^\top M_{\pi,\alpha}(s) r(s) \nonumber \\
& \qquad \qquad \qquad + |g_\pi(s)|^\top r(s)  \Big] \leq (1-\gamma)\beta, \\
& \quad 0 \leq r_i(s) \leq r_{\max}, \ \ \forall s \in \mathcal{S}, \ i=1,\dots,m.
\end{align}

\subsection{Derive a Closed-Form Solution}
\label{sec:appendix-derive-closed-form-solution}
The Lagrangian is constructed with $\lambda\geq 0$, $\eta(s) \succeq 0$ and $\xi(s) \succeq$ 0.
\begin{align}
& \quad \mathcal{L}(r,\lambda,\eta,\xi) = -\mathbb{E}_{s\sim d^\pi} [\textbf{1}^\top r(s)] + \lambda \Big( \mathbb{E}_{s\sim d^{\pi}}\Big[ \nonumber \\
&\qquad \frac{1}{2}r(s)^\top M_{\pi,\alpha}(s) r(s) + |g_\pi(s)|^\top r(s)\Big] - (1-\gamma)\beta \Big) \nonumber \\
&\qquad  - \mathbb{E}_{s\sim d^\pi} [ \eta(s)^\top r(s) ] + \mathbb{E}_{s\sim d^\pi} [ \xi(s)^\top (r(s) - r_{\max}) ] .
\end{align}
When the optimal solution $r^*_i(s)$ satisfies $0<r^*_i(s)<r_{\max}$ for every state dimension $i$, by complementary slackness we have
\begin{gather}
    \eta(s)^\top r^*(s)  = 0 \\
      \xi(s)^\top (r^*(s) - r_{\max})  = 0
\end{gather}
\begin{align}
& \quad \lambda \Big( \mathbb{E}_{s\sim d^{\pi}}\Big[ \frac{1}{2}r^*(s)^\top M_{\pi,\alpha}(s) r^*(s) \nonumber \\
&\qquad + |g_\pi(s)|^\top r^*(s)\Big] - (1-\gamma)\beta \Big) = 0 \label{eq:complementary-slackness-3}
\end{align}
for all sampled states and each state dimension $i$. Since $0<r^*_i(s)<r_{\max}$ , the only solution is that $\eta(s) = \xi(s) = 0$. Then we differentiate with respect to $r(s)$
\begin{equation}
    \frac{\partial \mathcal{L}}{\partial r(s)} = -\textbf{1} + \lambda ( M_{\pi,\alpha}(s)r(s) + |g_\pi(s)|).
\end{equation}
For the optimal solution $r^*(s)$ we have this gradient equals 0, meaning
\begin{equation}
    -\textbf{1} + \lambda ( M_{\pi,\alpha}(s)r^*(s) + |g_\pi(s)|) = 0
\end{equation}
Since $M_{\pi,\alpha}(s) \succ 0$, $r^*_i(s) > 0$ and $|g_\pi(s)| \succeq 0$, $\lambda=0$ does not allow this to happen. Therefore, $\lambda>0$ holds and we 
\begin{equation}
  r^*(s) = M_{\pi,\alpha}(s)^{-1} (\frac{1}{\lambda} \textbf{1} - |g_{\pi}(s)|),\label{eq:optimal-solution-with-lambda} 
\end{equation}
and Eq. \eqref{eq:complementary-slackness-3} is equivalent to
\begin{equation}
    \mathbb{E}_{s\sim d^{\pi}}\Big[ \frac{1}{2}r^*(s)^\top M_{\pi,\alpha}(s) r^*(s) + |g_\pi(s)|^\top r^*(s)\Big] = (1-\gamma)\beta. \label{eq:complementary-slackness-with-positive-lambda}
\end{equation}
Solving $r^*(s)$ in Eq. \eqref{eq:optimal-solution-with-lambda} and \eqref{eq:complementary-slackness-with-positive-lambda} derives our closed-form solution 
\begin{equation}
r^*(s)=M_{\pi,\alpha}(s)^{-1} \left(\kappa \textbf{1} - |g_\pi(s)| \right), \label{eq:nuus-final-solution}
\end{equation}
where
\begin{equation}
\kappa= \sqrt{\frac{2(1-\gamma)\beta + \mathbb{E}_{s'\sim d^\pi} \left[ |g_\pi(s')|^\top M_{\pi,\alpha}(s')^{-1}|g_\pi(s')| \right]}{\mathbb{E}_{s'\sim d^\pi} \left[ \textbf{1}^\top M_{\pi,\alpha}(s')^{-1}\textbf{1} \right]}}.
\label{eq:nuus-solution-kappa-term}
\end{equation}
However, the closed-form solution is based on $0<r_i(s)<r_{\max}$, and we have to calibrate $\alpha$ to guarantee that 
\begin{equation}
    0 < M_{\pi,\alpha}(s)^{-1} \left(\kappa \textbf{1} - |g_\pi(s)| \right) < r_{\max}.
\end{equation}

\section{Additional Implementation Details}
\label{sec:appendix-additional-implementation-details}

\subsection{Exploit Gaussian Policy for \method{}}

In our experiments, the policy is parameterized as a Gaussian distribution $\pi(\cdot|s) = \mathcal{N}(\mu(s), \Sigma)$ where $\mu(s)$ is a neural network for predicting the mean, and $\Sigma=\text{diag}(\sigma_1^2,\dots,\sigma^2_d)$ is a learnable but state-independent diagonal covariance matrix. The terms $H_{\pi,\alpha}(s)$ and $g_\pi(s)$ in Eq. \eqref{eq:nuus-reformulation-H-matrix} and \eqref{eq:nuus-g_pi} involve $\nabla_s \log\pi(a|s)$, $\nabla^2_s \log\pi(a|s)$, and $A^\pi(s,a)$ in the expectation terms. We leverage the structure of Gaussian policy to avoid explicitly estimating $\nabla_s \log \pi(a|s)$ and $\nabla^2_s \log \pi(a|s)$. Let $J_\mu(s) = \frac{\partial \mu(s)}{\partial s}$ denote the Jacobian matrix of the policy mean w.r.t. the state, our implementation uses
\begin{gather}
    g_\pi(s) = J_{\mu}(s)^\top \mathbb{E}_{a\sim \pi} [\nabla_a A^\pi(s,a)]
\end{gather}
and
\begin{align}
& \quad H_{\pi,\alpha}(s) \approx \alpha J_{\mu}(s)^\top \Sigma^{-1} J_{\mu}(s) \nonumber \\
&\qquad - J_{\mu}(s)^\top \mathbb{E}_{a\sim \pi}[\nabla_a^2 A^\pi(s,a)]J_{\mu}(s)
\end{align}

\subsection{Experiment Setup}
All experiments were conducted on Quadro RTX 6000 GPU. We build on the official implementation of SA-PPO, ATLA-PPO and WocaR-PPO. For all methods, the actor and critic are parameterized by multilayer perceptrons with hidden layers $[64,64]$, and each policy is trained for $2\times 10^6$ environment steps. We made two changes compared to the official implementation. First, we train an auxiliary advantage network $A^\pi(s,a)$ of hidden layers $[64,64]$ alongside the policy. Second, while the official implementation enables reward normalization, we disable it because naively reversing reward normalization on the learned advantage network does not provide reliable advantage estimates. We train each algorithm using four random seeds. 

We train each PA-AD attacker for $1\times 10^6$ environment steps. Its actor and critic use the same MLP architecture, with hidden layers $[64, 64]$, and are both optimized with a learning rate of $3\times10^{-4}$.


\section{Additional Experimental Results}
\label{sec:appendix-additional-experimental-results}

Table \ref{tab:hopper-discounted-return-degradation}, Table \ref{tab:halfcheetah-discounted-return-degradation} and Table \ref{tab:walker2d-discounted-return-degradation} present the discounted total return and the degradation compared to the clean policy under Random, Critic, MAD and PA-AD attackers on Hopper, HalfCheetah and Walker2d environments, respectively. We use these results to generate Table 4 in the main paper by taking the largest degradation across all state attackers.

In Table \ref{tab:nuus-alpha-ablation} we present the degradation in discounted return caused by different $\alpha$ values, and the percentage of perturbation budget $r(s)$ reaching the upper bound $r_{\max}$.

Table \ref{tab:pd-nuus-kappa} reports the empirical values of global multiplier $\kappa$ for PD-NUUS.

Table \ref{tab:nuus-paad-rewards} reports the episode rewards of clean policies and policies perturbed by PD-NUUS ($\beta\in\{20,40\}$) and PA-AD attacker.

\begin{figure*}[t]
    \centering
    \includegraphics[width=0.95\linewidth]{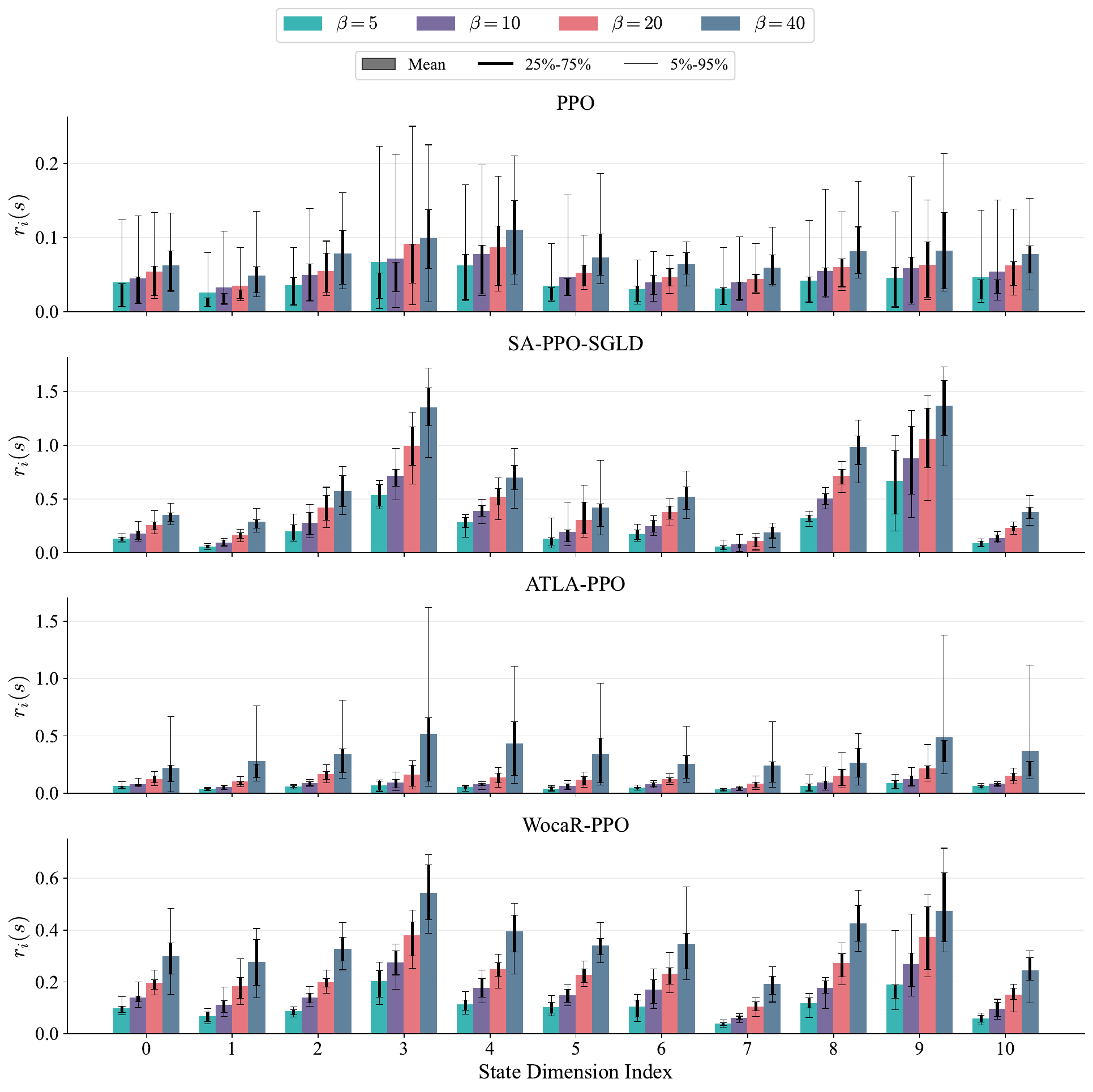}
    \caption{The mean and confidence range of calculated budget $r_i(s)$ of policies under PA-AD attack sampled from \method{} with $\beta \in \{ 5,10,20,40 \}$ on Hopper. SA-PPO-SGLD, ATLA-PPO and WocaR-PPO admit higher perturbation tolerance than PPO in nearly all state dimensions.}.
    \label{fig:nuus-budgets-all-betas}
\end{figure*}

In Figure \ref{fig:nuus-budgets-all-betas} we show the visualized budgets of \method{} for PPO, SA-PPO-SGLD, ATLA-PPO and WocaR-PPO policies when $\beta \in \{ 5, 10, 20, 40\}$.

\begin{table*}[t]
\centering

\renewcommand{\arraystretch}{0.95}
\fontsize{9pt}{11pt}\selectfont
\setlength{\tabcolsep}{5pt}

\begin{tabular}{cclcccc}
    \toprule
    \multirow{2}{*}{\textbf{Set}}
    & \multirow{2}{*}{\textbf{Parameter}}
    & \multirow{2}{*}{\textbf{Attacker}}
    & \multicolumn{4}{c}{\textbf{Algorithm}} \\
    \cmidrule(lr){4-7}
    & & & PPO & SA-PPO-SGLD & ATLA-PPO & WocaR-PPO \\
    \midrule

    No Attack & -- & --
    & 269.2 $\pm$ 10.0
    & 256.4 $\pm$ 6.5
    & 257.5 $\pm$ 6.2
    & 267.9 $\pm$ 4.5 \\

    \midrule

    \multirow{16}{*}{$L_\infty$-Norm}
    & \multirow{4}{*}{$\epsilon=0.05$}
    & Random
    & 268.2 $\pm$ 9.1 (1.0)
    & 256.1 $\pm$ 6.3 (0.3)
    & 257.2 $\pm$ 5.8 (0.3)
    & 268.0 $\pm$ 4.3 (-0.1) \\
    &
    & Critic
    & 269.0 $\pm$ 11.0 (0.2)
    & 255.9 $\pm$ 6.2 (0.5)
    & 258.3 $\pm$ 5.7 (-0.8)
    & 268.4 $\pm$ 5.7 (-0.5) \\
    &
    & MAD
    & 265.9 $\pm$ 9.1 (3.3)
    & 256.1 $\pm$ 5.5 (0.3)
    & 254.8 $\pm$ 6.5 (2.7)
    & 267.4 $\pm$ 4.3 (0.5) \\
    &
    & PA-AD
    & \textbf{217.2 $\pm$ 29.3 (52.0)}
    & \textbf{255.8 $\pm$ 4.8 (0.6)}
    & \textbf{240.2 $\pm$ 17.7 (17.3)}
    & \textbf{266.7 $\pm$ 2.1 (1.2)} \\

    \cmidrule(lr){2-7}

    &
    \multirow{4}{*}{$\epsilon=0.10$}
    & Random
    & 267.5 $\pm$ 8.8 (1.7)
    & 256.0 $\pm$ 6.3 (0.4)
    & 256.3 $\pm$ 6.5 (1.2)
    & 267.8 $\pm$ 4.3 (0.1) \\
    &
    & Critic
    & 266.0 $\pm$ 11.2 (3.2)
    & 255.9 $\pm$ 5.7 (0.5)
    & 252.7 $\pm$ 8.3 (4.8)
    & 270.7 $\pm$ 3.5 (-2.8) \\
    &
    & MAD
    & 241.4 $\pm$ 14.9 (27.8)
    & \textbf{255.8 $\pm$ 5.5 (0.6)}
    & 243.5 $\pm$ 9.7 (14.0)
    & 265.4 $\pm$ 4.2 (2.5) \\
    &
    & PA-AD
    & \textbf{106.9 $\pm$ 64.9 (162.3)}
    & 260.4 $\pm$ 7.4 (-4.0)
    & \textbf{178.8 $\pm$ 60.7 (78.7)}
    & \textbf{257.3 $\pm$ 8.1 (10.6)} \\

    \cmidrule(lr){2-7}

    &
    \multirow{4}{*}{$\epsilon=0.15$}
    & Random
    & 264.0 $\pm$ 8.3 (5.2)
    & 256.1 $\pm$ 5.8 (0.3)
    & 256.4 $\pm$ 6.3 (1.1)
    & 267.6 $\pm$ 4.0 (0.3) \\
    &
    & Critic
    & 260.9 $\pm$ 11.6 (8.3)
    & \textbf{255.8 $\pm$ 5.8 (0.6)}
    & 237.5 $\pm$ 28.0 (20.0)
    & 270.7 $\pm$ 3.4 (-2.8) \\
    &
    & MAD
    & 186.0 $\pm$ 46.0 (83.2)
    & 257.0 $\pm$ 4.9 (-0.6)
    & 230.8 $\pm$ 15.9 (26.7)
    & 262.6 $\pm$ 3.4 (5.3) \\
    &
    & PA-AD
    & \textbf{100.3 $\pm$ 66.5 (168.9)}
    & 256.2 $\pm$ 13.1 (0.2)
    & \textbf{154.5 $\pm$ 69.2 (103.0)}
    & \textbf{242.2 $\pm$ 16.2 (25.7)} \\

    \cmidrule(lr){2-7}

    &
    \multirow{4}{*}{$\epsilon=0.20$}
    & Random
    & 263.7 $\pm$ 7.8 (5.5)
    & 255.7 $\pm$ 6.4 (0.7)
    & 252.6 $\pm$ 7.4 (4.9)
    & 266.6 $\pm$ 4.5 (1.3) \\
    &
    & Critic
    & 242.5 $\pm$ 25.8 (26.7)
    & 255.0 $\pm$ 6.6 (1.4)
    & 233.5 $\pm$ 26.3 (24.0)
    & 270.0 $\pm$ 3.5 (-2.1) \\
    &
    & MAD
    & 162.8 $\pm$ 54.2 (106.4)
    & 255.2 $\pm$ 3.0 (1.2)
    & 223.1 $\pm$ 15.1 (34.4)
    & 258.0 $\pm$ 6.4 (9.9) \\
    &
    & PA-AD
    & \textbf{70.7 $\pm$ 34.0 (198.5)}
    & \textbf{228.0 $\pm$ 29.4 (28.4)}
    & \textbf{120.7 $\pm$ 64.8 (136.8)}
    & \textbf{116.2 $\pm$ 43.2 (151.7)} \\

    \midrule

    \multirow{16}{*}{PD-NUUS}
    & \multirow{4}{*}{$\beta=5$}
    & Random
    & 268.9 $\pm$ 10.3 (0.3)
    & 255.5 $\pm$ 5.6 (0.9)
    & 257.3 $\pm$ 5.8 (0.2)
    & 268.0 $\pm$ 4.2 (-0.1) \\
    &
    & Critic
    & 268.6 $\pm$ 11.2 (0.6)
    & 253.6 $\pm$ 6.3 (2.8)
    & 257.3 $\pm$ 5.4 (0.2)
    & \textbf{265.2 $\pm$ 7.5 (2.7)} \\
    &
    & MAD
    & \textbf{267.5 $\pm$ 10.3 (1.7)}
    & \textbf{252.5 $\pm$ 8.8 (3.9)}
    & \textbf{256.5 $\pm$ 5.6 (1.0)}
    & 267.4 $\pm$ 4.6 (0.5) \\
    &
    & PA-AD
    & 268.5 $\pm$ 10.8 (0.7)
    & 257.4 $\pm$ 6.6 (-1.0)
    & 256.6 $\pm$ 6.3 (0.9)
    & 266.5 $\pm$ 5.9 (1.4) \\

    \cmidrule(lr){2-7}

    &
    \multirow{4}{*}{$\beta=10$}
    & Random
    & 268.9 $\pm$ 10.3 (0.3)
    & 256.1 $\pm$ 5.5 (0.3)
    & 257.3 $\pm$ 6.0 (0.2)
    & 267.3 $\pm$ 4.8 (0.6) \\
    &
    & Critic
    & 268.3 $\pm$ 11.6 (0.9)
    & 252.3 $\pm$ 7.4 (4.1)
    & 256.7 $\pm$ 6.1 (0.8)
    & \textbf{265.5 $\pm$ 6.0 (2.4)} \\
    &
    & MAD
    & 267.2 $\pm$ 9.5 (2.0)
    & \textbf{250.6 $\pm$ 12.0 (5.8)}
    & \textbf{255.0 $\pm$ 6.0 (2.5)}
    & 267.1 $\pm$ 4.6 (0.8) \\
    &
    & PA-AD
    & \textbf{259.2 $\pm$ 17.5 (10.0)}
    & 251.5 $\pm$ 5.6 (4.9)
    & 257.1 $\pm$ 5.6 (0.4)
    & 266.0 $\pm$ 6.5 (1.9) \\

    \cmidrule(lr){2-7}

    &
    \multirow{4}{*}{$\beta=20$}
    & Random
    & 268.5 $\pm$ 9.5 (0.7)
    & 255.6 $\pm$ 6.4 (0.8)
    & 256.1 $\pm$ 6.7 (1.4)
    & 267.3 $\pm$ 4.6 (0.6) \\
    &
    & Critic
    & 269.0 $\pm$ 11.6 (0.2)
    & 250.6 $\pm$ 8.3 (5.8)
    & 257.2 $\pm$ 6.0 (0.3)
    & 262.7 $\pm$ 10.9 (5.2) \\
    &
    & MAD
    & \textbf{266.5 $\pm$ 9.4 (2.7)}
    & \textbf{241.8 $\pm$ 20.3 (14.6)}
    & 249.7 $\pm$ 9.4 (7.8)
    & 259.1 $\pm$ 16.1 (8.8) \\
    &
    & PA-AD
    & 268.4 $\pm$ 11.9 (0.8)
    & 249.2 $\pm$ 11.5 (7.2)
    & \textbf{233.0 $\pm$ 23.6 (24.5)}
    & \textbf{258.6 $\pm$ 13.7 (9.3)} \\

    \cmidrule(lr){2-7}

    &
    \multirow{4}{*}{$\beta=40$}
    & Random
    & 268.4 $\pm$ 9.5 (0.8)
    & 254.5 $\pm$ 6.1 (1.9)
    & 257.2 $\pm$ 5.3 (0.3)
    & 266.6 $\pm$ 4.7 (1.3) \\
    &
    & Critic
    & 267.6 $\pm$ 12.2 (1.6)
    & 248.2 $\pm$ 5.8 (8.2)
    & 252.4 $\pm$ 7.5 (5.1)
    & 261.4 $\pm$ 11.9 (6.5) \\
    &
    & MAD
    & 261.3 $\pm$ 12.9 (7.9)
    & \textbf{225.6 $\pm$ 27.0 (30.8)}
    & 238.8 $\pm$ 17.3 (18.7)
    & 253.5 $\pm$ 22.8 (14.4) \\
    &
    & PA-AD
    & \textbf{248.2 $\pm$ 16.6 (21.0)}
    & 226.7 $\pm$ 12.4 (29.7)
    & \textbf{129.8 $\pm$ 95.9 (127.7)}
    & \textbf{246.2 $\pm$ 8.6 (21.7)} \\

    \bottomrule
\end{tabular}
\caption{Discounted returns ($\gamma=0.99$) under state attackers on Hopper environment. When using \method{}, the values of $\alpha$ for PPO, SA-PPO-SGLD, ATLA-PPO and WocaR-PPO are 0.15, 0.25, 0.0075 and 0.2, respectively. The values in parentheses represent degradation compared to no attack.}
\label{tab:hopper-discounted-return-degradation}
\end{table*}

\begin{table*}[t]
\centering
\renewcommand{\arraystretch}{0.95}
\fontsize{9pt}{11pt}\selectfont
\setlength{\tabcolsep}{5pt}
\begin{tabular}{cclcccc}
    \toprule
    \multirow{2}{*}{\textbf{Set}}
    & \multirow{2}{*}{\textbf{Parameter}}
    & \multirow{2}{*}{\textbf{Attacker}}
    & \multicolumn{4}{c}{\textbf{Algorithm}} \\
    \cmidrule(lr){4-7}
    & & & PPO & SA-PPO-SGLD & ATLA-PPO & WocaR-PPO \\
    \midrule

    No Attack & -- & --
    & 421.5 $\pm$ 57.5
    & 349.4 $\pm$ 40.3
    & 345.7 $\pm$ 24.9
    & 331.4 $\pm$ 10.4 \\

    \midrule

    \multirow{16}{*}{$L_\infty$-Norm}
    & \multirow{4}{*}{$\epsilon=0.05$}
    & Random
    & 415.0 $\pm$ 46.7 (6.5)
    & 350.5 $\pm$ 40.7 (-1.1)
    & 342.9 $\pm$ 21.0 (2.8)
    & 331.3 $\pm$ 10.1 (0.1) \\
    &
    & Critic
    & 404.7 $\pm$ 72.6 (16.8)
    & 350.4 $\pm$ 39.3 (-1.0)
    & 344.3 $\pm$ 18.8 (1.4)
    & 331.6 $\pm$ 10.7 (-0.2) \\
    &
    & MAD
    & \textbf{336.7 $\pm$ 47.7 (84.8)}
    & 349.9 $\pm$ 37.5 (-0.5)
    & 339.1 $\pm$ 21.4 (6.6)
    & 328.4 $\pm$ 9.3 (3.0) \\
    &
    & PA-AD
    & 338.3 $\pm$ 67.0 (83.2)
    & \textbf{348.8 $\pm$ 38.6 (0.6)}
    & \textbf{313.1 $\pm$ 38.0 (32.6)}
    & \textbf{324.8 $\pm$ 7.1 (6.6)} \\

    \cmidrule(lr){2-7}

    &
    \multirow{4}{*}{$\epsilon=0.10$}
    & Random
    & 382.9 $\pm$ 44.2 (38.6)
    & 350.6 $\pm$ 37.9 (-1.2)
    & 345.2 $\pm$ 19.7 (0.5)
    & 328.2 $\pm$ 9.0 (3.2) \\
    &
    & Critic
    & 367.3 $\pm$ 98.8 (54.2)
    & 350.5 $\pm$ 41.5 (-1.1)
    & 334.1 $\pm$ 17.5 (11.6)
    & 329.5 $\pm$ 8.1 (1.9) \\
    &
    & MAD
    & 224.6 $\pm$ 58.6 (196.9)
    & 348.2 $\pm$ 40.0 (1.2)
    & 293.0 $\pm$ 42.8 (52.7)
    & \textbf{322.6 $\pm$ 7.0 (8.8)} \\
    &
    & PA-AD
    & \textbf{168.0 $\pm$ 59.8 (253.5)}
    & \textbf{348.0 $\pm$ 42.6 (1.4)}
    & \textbf{243.7 $\pm$ 44.2 (102.0)}
    & 328.4 $\pm$ 8.0 (3.0) \\

    \cmidrule(lr){2-7}

    &
    \multirow{4}{*}{$\epsilon=0.15$}
    & Random
    & 368.3 $\pm$ 36.5 (53.2)
    & 351.2 $\pm$ 38.2 (-1.8)
    & 339.6 $\pm$ 24.5 (6.1)
    & 326.7 $\pm$ 9.8 (4.7) \\
    &
    & Critic
    & 332.2 $\pm$ 92.5 (89.3)
    & 351.1 $\pm$ 39.7 (-1.7)
    & 321.5 $\pm$ 15.7 (24.2)
    & 329.5 $\pm$ 10.1 (1.9) \\
    &
    & MAD
    & 162.1 $\pm$ 55.2 (259.4)
    & 344.6 $\pm$ 37.6 (4.8)
    & 232.8 $\pm$ 39.9 (112.9)
    & 315.3 $\pm$ 8.4 (16.1) \\
    &
    & PA-AD
    & \textbf{100.3 $\pm$ 47.7 (321.2)}
    & \textbf{339.7 $\pm$ 27.3 (9.7)}
    & \textbf{173.8 $\pm$ 53.7 (171.9)}
    & \textbf{300.9 $\pm$ 17.2 (30.5)} \\

    \cmidrule(lr){2-7}

    &
    \multirow{4}{*}{$\epsilon=0.20$}
    & Random
    & 321.8 $\pm$ 38.5 (99.7)
    & 348.5 $\pm$ 40.1 (0.9)
    & 328.3 $\pm$ 26.3 (17.4)
    & 324.3 $\pm$ 8.5 (7.1) \\
    &
    & Critic
    & 280.5 $\pm$ 100.5 (141.0)
    & 347.2 $\pm$ 42.7 (2.2)
    & 295.5 $\pm$ 16.0 (50.2)
    & 324.7 $\pm$ 12.2 (6.7) \\
    &
    & MAD
    & 117.1 $\pm$ 42.6 (304.4)
    & 342.5 $\pm$ 38.2 (6.9)
    & 164.5 $\pm$ 36.9 (181.2)
    & 303.2 $\pm$ 6.8 (28.2) \\
    &
    & PA-AD
    & \textbf{22.3 $\pm$ 33.6 (399.2)}
    & \textbf{337.8 $\pm$ 42.7 (11.6)}
    & \textbf{86.6 $\pm$ 55.7 (259.1)}
    & \textbf{287.0 $\pm$ 12.2 (44.4)} \\

    \midrule

    \multirow{16}{*}{PD-NUUS}
    & \multirow{4}{*}{$\beta=5$}
    & Random
    & 421.2 $\pm$ 60.9 (0.3)
    & 350.0 $\pm$ 39.9 (-0.6)
    & 349.8 $\pm$ 22.7 (-4.1)
    & 330.1 $\pm$ 11.9 (1.3) \\
    &
    & Critic
    & 418.1 $\pm$ 56.6 (3.4)
    & 350.4 $\pm$ 41.3 (-1.0)
    & 350.2 $\pm$ 20.1 (-4.5)
    & 332.1 $\pm$ 9.4 (-0.7) \\
    &
    & MAD
    & \textbf{409.2 $\pm$ 65.3 (12.3)}
    & 349.6 $\pm$ 41.4 (-0.2)
    & 349.4 $\pm$ 20.0 (-3.7)
    & \textbf{329.6 $\pm$ 9.2 (1.8)} \\
    &
    & PA-AD
    & 418.2 $\pm$ 60.8 (3.3)
    & \textbf{346.2 $\pm$ 41.2 (3.2)}
    & \textbf{347.9 $\pm$ 21.7 (-2.2)}
    & 329.9 $\pm$ 8.9 (1.5) \\
    
    \cmidrule(lr){2-7}
    
    &
    \multirow{4}{*}{$\beta=10$}
    & Random
    & \textbf{417.6 $\pm$ 58.1 (3.9)}
    & 348.3 $\pm$ 41.2 (1.1)
    & 350.1 $\pm$ 19.2 (-4.4)
    & 330.4 $\pm$ 9.6 (1.0) \\
    &
    & Critic
    & 417.9 $\pm$ 59.9 (3.6)
    & 351.5 $\pm$ 40.4 (-2.1)
    & 351.4 $\pm$ 18.3 (-5.7)
    & 333.4 $\pm$ 10.3 (-2.0) \\
    &
    & MAD
    & 422.4 $\pm$ 60.0 (-0.9)
    & \textbf{348.1 $\pm$ 38.8 (1.3)}
    & 346.3 $\pm$ 23.8 (-0.6)
    & 329.2 $\pm$ 11.4 (2.2) \\
    &
    & PA-AD
    & 424.4 $\pm$ 58.1 (-2.9)
    & 348.6 $\pm$ 38.8 (0.8)
    & \textbf{344.2 $\pm$ 21.2 (1.5)}
    & \textbf{325.6 $\pm$ 8.6 (5.8)} \\
    
    \cmidrule(lr){2-7}
    
    &
    \multirow{4}{*}{$\beta=20$}
    & Random
    & 409.3 $\pm$ 59.1 (12.2)
    & 347.7 $\pm$ 39.3 (1.7)
    & 350.3 $\pm$ 21.2 (-4.6)
    & 329.8 $\pm$ 10.2 (1.6) \\
    &
    & Critic
    & 414.8 $\pm$ 65.6 (6.7)
    & 350.5 $\pm$ 40.4 (-1.1)
    & 346.1 $\pm$ 20.4 (-0.4)
    & 331.4 $\pm$ 11.2 (0.0) \\
    &
    & MAD
    & 413.9 $\pm$ 68.6 (7.6)
    & 346.5 $\pm$ 36.8 (2.9)
    & 344.1 $\pm$ 24.5 (1.6)
    & 326.8 $\pm$ 12.2 (4.6) \\
    &
    & PA-AD
    & \textbf{404.3 $\pm$ 54.5 (17.2)}
    & \textbf{345.2 $\pm$ 38.7 (4.2)}
    & \textbf{332.3 $\pm$ 19.0 (13.4)}
    & \textbf{311.5 $\pm$ 13.6 (19.9)} \\
    
    \cmidrule(lr){2-7}
    
    &
    \multirow{4}{*}{$\beta=40$}
    & Random
    & 416.7 $\pm$ 59.0 (4.8)
    & 347.1 $\pm$ 38.1 (2.3)
    & 349.6 $\pm$ 20.7 (-3.9)
    & 321.3 $\pm$ 12.8 (10.1) \\
    &
    & Critic
    & 413.6 $\pm$ 60.6 (7.9)
    & 340.4 $\pm$ 36.2 (9.0)
    & 335.2 $\pm$ 23.4 (10.5)
    & 324.0 $\pm$ 9.1 (7.4) \\
    &
    & MAD
    & 405.4 $\pm$ 61.6 (16.1)
    & 330.2 $\pm$ 29.4 (19.2)
    & 335.8 $\pm$ 24.6 (9.9)
    & 310.2 $\pm$ 11.9 (21.2) \\
    &
    & PA-AD
    & \textbf{387.3 $\pm$ 68.3 (34.2)}
    & \textbf{327.0 $\pm$ 44.3 (22.4)}
    & \textbf{319.6 $\pm$ 19.1 (26.1)}
    & \textbf{287.0 $\pm$ 24.8 (44.4)} \\

    \bottomrule
\end{tabular}
\caption{Discounted returns ($\gamma=0.99$) under state attackers on HalfCheetah environment. When using \method{}, the value of $\alpha$ for PPO, SA-PPO-SGLD, ATLA-PPO and WocaR-PPO are 0.06, 0.2, 0.02 and 0.075, respectively. The values in parentheses represent degradation compared to no attack.
}
\label{tab:halfcheetah-discounted-return-degradation}
\end{table*}

\begin{table*}[t]
\centering

\renewcommand{\arraystretch}{0.95}
\fontsize{9pt}{11pt}\selectfont
\setlength{\tabcolsep}{5pt}

\begin{tabular}{cclcccc}
    \toprule

    \multirow{2}{*}{\textbf{Set}}
    & \multirow{2}{*}{\textbf{Parameter}}
    & \multirow{2}{*}{\textbf{Attacker}}
    & \multicolumn{4}{c}{\textbf{Algorithm}} \\
    \cmidrule(lr){4-7}

    & & & PPO & SA-PPO-SGLD & ATLA-PPO & WocaR-PPO \\
    \midrule

    No Attack & -- & --
    & 278.5 $\pm$ 8.5
    & 298.3 $\pm$ 12.6
    & 276.7 $\pm$ 14.0
    & 287.9 $\pm$ 17.6
    \\

    \midrule

    \multirow{16}{*}{$L_\infty$-Norm}
    &
    \multirow{4}{*}{$\epsilon=0.05$}
    & Random
    & 278.2 $\pm$ 9.6 (0.3)
    & 299.0 $\pm$ 12.7 (-0.7)
    & 278.6 $\pm$ 11.5 (-1.9)
    & 287.6 $\pm$ 17.8 (0.3)
    \\
    &
    & Critic
    & 275.9 $\pm$ 9.4 (2.6)
    & 302.6 $\pm$ 10.8 (-4.3)
    & 275.6 $\pm$ 16.3 (1.1)
    & 289.5 $\pm$ 19.0 (-1.6)
    \\
    &
    & MAD
    & 265.1 $\pm$ 14.5 (13.4)
    & 299.6 $\pm$ 10.4 (-1.3)
    & 260.3 $\pm$ 10.4 (16.4)
    & 286.9 $\pm$ 17.6 (1.0)
    \\
    &
    & PA-AD
    & \textbf{228.2 $\pm$ 44.6 (50.3)}
    & \textbf{297.3 $\pm$ 14.4 (1.0)}
    & \textbf{227.7 $\pm$ 35.1 (49.0)}
    & \textbf{285.0 $\pm$ 19.0 (2.9)}
    \\

    \cmidrule(lr){2-7}

    &
    \multirow{4}{*}{$\epsilon=0.10$}
    & Random
    & 277.7 $\pm$ 9.7 (0.8)
    & 298.2 $\pm$ 12.8 (0.1)
    & 274.8 $\pm$ 12.2 (1.9)
    & 287.6 $\pm$ 17.8 (0.3)
    \\
    &
    & Critic
    & 260.5 $\pm$ 21.8 (18.0)
    & 296.6 $\pm$ 9.7 (1.7)
    & 255.0 $\pm$ 23.7 (21.7)
    & 292.3 $\pm$ 18.8 (-4.4)
    \\
    &
    & MAD
    & 229.9 $\pm$ 29.4 (48.6)
    & 295.3 $\pm$ 11.1 (3.0)
    & 222.1 $\pm$ 14.4 (54.6)
    & 282.8 $\pm$ 17.3 (5.1)
    \\
    &
    & PA-AD
    & \textbf{139.5 $\pm$ 16.5 (139.0)}
    & \textbf{275.3 $\pm$ 42.0 (23.0)}
    & \textbf{151.0 $\pm$ 17.6 (125.7)}
    & \textbf{245.6 $\pm$ 38.1 (42.3)}
    \\

    \cmidrule(lr){2-7}

    &
    \multirow{4}{*}{$\epsilon=0.15$}
    & Random
    & 268.5 $\pm$ 13.3 (10.0)
    & 298.1 $\pm$ 11.7 (0.2)
    & 268.9 $\pm$ 11.7 (7.8)
    & 286.7 $\pm$ 18.4 (1.2)
    \\
    &
    & Critic
    & 246.6 $\pm$ 25.1 (31.9)
    & 287.1 $\pm$ 28.2 (11.2)
    & 232.9 $\pm$ 29.7 (43.8)
    & 294.5 $\pm$ 17.8 (-6.6)
    \\
    &
    & MAD
    & 197.4 $\pm$ 23.0 (81.1)
    & 290.8 $\pm$ 8.6 (7.5)
    & 186.4 $\pm$ 19.4 (90.3)
    & 275.3 $\pm$ 17.2 (12.6)
    \\
    &
    & PA-AD
    & \textbf{116.2 $\pm$ 31.7 (162.3)}
    & \textbf{248.0 $\pm$ 17.9 (50.3)}
    & \textbf{111.6 $\pm$ 23.0 (165.1)}
    & \textbf{152.7 $\pm$ 60.8 (135.2)}
    \\

    \cmidrule(lr){2-7}

    &
    \multirow{4}{*}{$\epsilon=0.20$}
    & Random
    & 267.1 $\pm$ 13.0 (11.4)
    & 297.9 $\pm$ 10.4 (0.4)
    & 264.0 $\pm$ 11.4 (12.7)
    & 286.9 $\pm$ 17.9 (1.0)
    \\
    &
    & Critic
    & 227.1 $\pm$ 40.0 (51.4)
    & 271.8 $\pm$ 40.7 (26.5)
    & 221.1 $\pm$ 27.2 (55.6)
    & 297.7 $\pm$ 13.8 (-9.8)
    \\
    &
    & MAD
    & 178.5 $\pm$ 17.6 (100.0)
    & 278.5 $\pm$ 10.1 (19.8)
    & 162.7 $\pm$ 21.9 (114.0)
    & 272.0 $\pm$ 19.9 (15.9)
    \\
    &
    & PA-AD
    & \textbf{73.2 $\pm$ 33.5 (205.3)}
    & \textbf{169.4 $\pm$ 23.7 (128.9)}
    & \textbf{85.2 $\pm$ 41.7 (191.5)}
    & \textbf{148.9 $\pm$ 57.7 (139.0)}
    \\

    \midrule

    \multirow{16}{*}{PD-NUUS}
    &
    \multirow{4}{*}{$\beta=5$}
    & Random
    & 279.2 $\pm$ 7.6 (-0.7)
    & 298.0 $\pm$ 14.5 (0.3)
    & 276.8 $\pm$ 14.2 (-0.1)
    & 287.6 $\pm$ 17.9 (0.3)
    \\
    &
    & Critic
    & 276.5 $\pm$ 8.0 (2.0)
    & 305.3 $\pm$ 9.8 (-7.0)
    & \textbf{276.6 $\pm$ 15.5 (0.1)}
    & 289.3 $\pm$ 18.4 (-1.4)
    \\
    &
    & MAD
    & 279.3 $\pm$ 8.5 (-0.8)
    & 296.4 $\pm$ 14.4 (1.9)
    & 276.8 $\pm$ 11.9 (-0.1)
    & 287.9 $\pm$ 17.3 (0.0)
    \\
    &
    & PA-AD
    & \textbf{275.7 $\pm$ 6.4 (2.8)}
    & \textbf{282.3 $\pm$ 17.3 (16.0)}
    & 277.8 $\pm$ 10.0 (-1.1)
    & \textbf{282.3 $\pm$ 19.8 (5.6)}
    \\

    \cmidrule(lr){2-7}

    &
    \multirow{4}{*}{$\beta=10$}
    & Random
    & 278.0 $\pm$ 8.9 (0.5)
    & 297.6 $\pm$ 14.8 (0.7)
    & 277.6 $\pm$ 13.7 (-0.9)
    & 288.0 $\pm$ 17.3 (-0.1)
    \\
    &
    & Critic
    & 274.8 $\pm$ 10.6 (3.7)
    & 304.1 $\pm$ 7.1 (-5.8)
    & 276.1 $\pm$ 14.7 (0.6)
    & 289.7 $\pm$ 18.3 (-1.8)
    \\
    &
    & MAD
    & 276.2 $\pm$ 7.7 (2.3)
    & 297.0 $\pm$ 12.8 (1.3)
    & 276.0 $\pm$ 12.8 (0.7)
    & 287.7 $\pm$ 17.3 (0.2)
    \\
    &
    & PA-AD
    & \textbf{270.8 $\pm$ 11.9 (7.7)}
    & \textbf{283.4 $\pm$ 16.2 (14.9)}
    & \textbf{273.5 $\pm$ 15.5 (3.2)}
    & \textbf{286.2 $\pm$ 20.7 (1.7)}
    \\

    \cmidrule(lr){2-7}

    &
    \multirow{4}{*}{$\beta=20$}
    & Random
    & 277.9 $\pm$ 10.1 (0.6)
    & 297.6 $\pm$ 12.5 (0.7)
    & 278.2 $\pm$ 11.7 (-1.5)
    & 288.0 $\pm$ 17.8 (-0.1)
    \\
    &
    & Critic
    & 268.9 $\pm$ 13.7 (9.6)
    & 296.1 $\pm$ 12.4 (2.2)
    & 276.2 $\pm$ 14.0 (0.5)
    & 289.9 $\pm$ 18.9 (-2.0)
    \\
    &
    & MAD
    & 269.8 $\pm$ 10.2 (8.7)
    & 296.6 $\pm$ 6.1 (1.7)
    & 277.7 $\pm$ 13.2 (-1.0)
    & 286.6 $\pm$ 18.3 (1.3)
    \\
    &
    & PA-AD
    & \textbf{218.0 $\pm$ 40.7 (60.5)}
    & \textbf{282.9 $\pm$ 15.6 (15.4)}
    & \textbf{268.8 $\pm$ 11.3 (7.9)}
    & \textbf{271.2 $\pm$ 26.6 (16.7)}
    \\

    \cmidrule(lr){2-7}

    &
    \multirow{4}{*}{$\beta=40$}
    & Random
    & 277.2 $\pm$ 10.0 (1.3)
    & 296.3 $\pm$ 8.4 (2.0)
    & 275.2 $\pm$ 11.8 (1.5)
    & 287.4 $\pm$ 17.8 (0.5)
    \\
    &
    & Critic
    & 263.7 $\pm$ 18.3 (14.8)
    & 288.0 $\pm$ 8.8 (10.3)
    & 274.0 $\pm$ 16.1 (2.7)
    & 292.2 $\pm$ 17.3 (-4.3)
    \\
    &
    & MAD
    & 255.6 $\pm$ 8.3 (22.9)
    & 280.5 $\pm$ 10.1 (17.8)
    & 272.1 $\pm$ 11.8 (4.6)
    & 282.6 $\pm$ 17.8 (5.3)
    \\
    &
    & PA-AD
    & \textbf{172.9 $\pm$ 44.8 (105.6)}
    & \textbf{170.0 $\pm$ 81.1 (128.3)}
    & \textbf{264.4 $\pm$ 22.6 (12.3)}
    & \textbf{237.9 $\pm$ 50.4 (50.0)}
    \\

    \bottomrule
\end{tabular}
\caption{Discounted returns ($\gamma=0.99$) under state attackers on Walker2d environment. When using \method{}, the value of $\alpha$ for PPO, SA-PPO-SGLD, ATLA-PPO and WocaR-PPO are 0.025, 0.075, 0.05 and 0.15, respectively. The values in parentheses represent degradation compared to no attack.
}
\label{tab:walker2d-discounted-return-degradation}
\end{table*}

\begin{table*}[t]
    \centering

    \begin{tabular}{llccc}
        \toprule
        \textbf{Environment}
        & \textbf{Policy}
        & $\alpha$
        & \textbf{Budget $r(s)$ at $r_{\max}$ ($\%$)}
        & \textbf{Degradation in Discounted Return} \\
        \midrule

        \multirow{12}{*}{Hopper}
        & \multirow{3}{*}{PPO}
        & 0    & 36.219 & 241.4 \\
        & & 0.015 & 4.855 & 234.7  \\
        & & 0.15${}^*$ & 0.137 & 21.0 \\
        \cmidrule(lr){2-5}

        & \multirow{2}{*}{SA-PPO-SGLD}
        & 0    & 18.343 & 72.3 \\
        & & 0.25${}^*$ & 7.515 & 29.6 \\
        & & 0.3 & 8.758 & 22.4 \\
        \cmidrule(lr){2-5}

        & \multirow{2}{*}{ATLA-PPO}
        & 0      & 15.982 & 176.7 \\
        & & 0.0075${}^*$ & 0.325 & 127.7 \\
        & & 0.02 & 0.033 & 17.7  \\
        \cmidrule(lr){2-5}

        & \multirow{3}{*}{WocaR-PPO}
        & 0   & 18.400 & 183.2 \\
        & & 0.15 & 4.021  & 189.0 \\
        & & 0.2${}^*$ & 1.145 & 21.7 \\











        \bottomrule
    \end{tabular}

    \caption{Percentage of perturbation budgets $r(s)$ that reach the upper bound $r_{\max}$ under PA-AD attacks with \method{} at $\beta=40$. We compare the selected value (with ${}^*$) of $\alpha$ with other $\alpha$ values. An inappropriate choice of $\alpha$ yields degradation in discounted total return substantially greater than $\beta$ and cause a larger fraction of the learned budgets to saturate at $r_{\max}$.
    }
    \label{tab:nuus-alpha-ablation}
\end{table*}

\begin{table*}[t]
    \centering
    \renewcommand{\arraystretch}{1.15}
    \fontsize{9pt}{11pt}\selectfont
    \setlength{\tabcolsep}{3pt}

    \begin{tabular}{lccccc}
        \toprule[1.2pt]
        \textbf{Environment} & $\beta$
        & PPO
        & SA-PPO-SGLD
        & ATLA-PPO
        & WocaR-PPO \\
        \midrule

        \multirow{4}{*}{Hopper}
        & 5  & $0.458 \pm 0.077$ & $0.026 \pm 0.001$
             & $0.136 \pm 0.024$ & $0.063 \pm 0.003$ \\
        & 10 & $0.643 \pm 0.109$ & $0.034 \pm 0.002$
             & $0.182 \pm 0.036$ & $0.086 \pm 0.004$ \\
        & 20 & $0.906 \pm 0.154$ & $0.045 \pm 0.002$
             & $0.249 \pm 0.052$ & $0.118 \pm 0.005$ \\
        & 40 & $1.279 \pm 0.218$ & $0.062 \pm 0.003$
             & $0.347 \pm 0.075$ & $0.166 \pm 0.007$ \\
        \midrule

        \multirow{4}{*}{HalfCheetah}
        & 5  & $0.155 \pm 0.029$ & $0.021 \pm 0.006$
             & $0.130 \pm 0.029$ & $0.015 \pm 0.003$ \\
        & 10 & $0.216 \pm 0.041$ & $0.028 \pm 0.007$
             & $0.180 \pm 0.041$ & $0.019 \pm 0.004$ \\
        & 20 & $0.304 \pm 0.058$ & $0.038 \pm 0.010$
             & $0.251 \pm 0.058$ & $0.026 \pm 0.005$ \\
        & 40 & $0.429 \pm 0.082$ & $0.053 \pm 0.014$
             & $0.353 \pm 0.082$ & $0.035 \pm 0.007$ \\
        \midrule

        \multirow{4}{*}{Walker2d}
        & 5  & $0.210 \pm 0.026$ & $0.051 \pm 0.002$
             & $0.338 \pm 0.045$ & $0.072 \pm 0.006$ \\
        & 10 & $0.282 \pm 0.037$ & $0.064 \pm 0.003$
             & $0.467 \pm 0.064$ & $0.095 \pm 0.008$ \\
        & 20 & $0.388 \pm 0.052$ & $0.085 \pm 0.004$
             & $0.653 \pm 0.092$ & $0.129 \pm 0.011$ \\
        & 40 & $0.541 \pm 0.074$ & $0.116 \pm 0.006$
             & $0.917 \pm 0.130$ & $0.178 \pm 0.014$ \\

        \bottomrule[1.2pt]
    \end{tabular}

    \caption{The mean and standard deviation of $\kappa$ values estimated from sampled trajectories, when synthesizing PD-NUUS.}
    \label{tab:pd-nuus-kappa}
\end{table*}

\begin{table*}[t]
    \centering
    \fontsize{9pt}{11pt}\selectfont
    \setlength{\tabcolsep}{4pt}
    \begin{tabular}{llccccc}
        \toprule
        \textbf{Environment} & \textbf{Attacker} & $\beta$
        & PPO & SA-PPO-SGLD & ATLA-PPO & WocaR-PPO \\
        \midrule
        \multirow{3}{*}{Hopper} & -- & --
        & $3471.7 \pm 391.7$
        & $3203.0 \pm 378.5$
        & $3227.5 \pm 634.9$
        & $3358.3 \pm 452.8$ \\
        & PD-NUUS & 20
        & $3112.9 \pm 757.8$
        & $1802.7 \pm 1311.8$
        & $1573.8 \pm 1111.0$
        & $2501.1 \pm 1173.0$ \\
        & PD-NUUS & 40
        & $2489.8 \pm 1287.3$
        & $1198.2 \pm 1085.3$
        & $924.3 \pm 1323.3$
        & $1663.1 \pm 1081.5$ \\
        \midrule
        \multirow{3}{*}{HalfCheetah} & -- & --
        & $5275.1 \pm 1586.0$
        & $4320.3 \pm 611.3$
        & $4267.1 \pm 854.5$
        & $3952.2 \pm 566.3$ \\
        & PD-NUUS & 20
        & $4793.0 \pm 1810.3$
        & $4241.0 \pm 561.7$
        & $3930.9 \pm 954.0$
        & $2339.6 \pm 1389.2$ \\
        & PD-NUUS & 40
        & $4356.1 \pm 1986.7$
        & $3952.1 \pm 751.8$
        & $3730.3 \pm 971.5$
        & $2165.6 \pm 1193.5$ \\
        \midrule
        \multirow{3}{*}{Walker2d} & -- & --
        & $3887.0 \pm 1290.8$
        & $3858.3 \pm 1244.0$
        & $3647.7 \pm 1241.3$
        & $4271.3 \pm 681.9$ \\
        & PD-NUUS & 20
        & $1713.6 \pm 1704.0$
        & $3426.0 \pm 1421.2$
        & $2872.9 \pm 1388.8$
        & $3213.1 \pm 1562.4$ \\
        & PD-NUUS & 40
        & $921.1 \pm 1333.2$
        & $1347.8 \pm 1251.3$
        & $2992.0 \pm 1453.8$
        & $2526.1 \pm 1526.7$ \\
        \bottomrule
    \end{tabular}
    \caption{Episode rewards $\mathbb{E}[\sum_{t=0}^T R(s_t,a_t)]$ without attack and under PA-AD attack with PD-NUUS ($\beta \in \{20,40\}$).}
    \label{tab:nuus-paad-rewards}
\end{table*}

\end{document}